\pdfoutput=1
\documentclass{article}

\usepackage[main, final]{neurips_2026}

\usepackage[utf8]{inputenc}
\usepackage[T1]{fontenc}
\usepackage{hyperref}
\usepackage{url}
\usepackage{booktabs}
\usepackage{amsfonts}
\usepackage{nicefrac}
\usepackage{microtype}
\usepackage{xcolor}

\usepackage{amsmath}
\usepackage{amssymb}
\usepackage{mathtools}
\usepackage{amsthm}

\theoremstyle{plain}

\theoremstyle{definition}

\theoremstyle{remark}

\usepackage{lineno}
\usepackage{xspace}
\usepackage{mathtools}

\definecolor{darkblue}{rgb}{0, 0, 0.5}
\hypersetup{colorlinks=true, citecolor=darkblue, linkcolor=darkblue, urlcolor=darkblue}

\usepackage{graphicx}
\usepackage{subcaption}
\usepackage[space]{grffile}
\usepackage{listings}
\usepackage{subcaption}
\usepackage{inconsolata}
\usepackage[inline]{enumitem}
\usepackage[font=small,labelfont=bf]{caption}
\usepackage{wrapfig}
\usepackage{enumitem}

\usepackage{multirow}

\definecolor{codegreen}{rgb}{0,0.6,0}
\definecolor{codegray}{rgb}{0.5,0.5,0.5}
\definecolor{codepurple}{rgb}{0.58,0,0.82}

\definecolor{backcolour}{RGB}{245,248,250}
\definecolor{emph}{RGB}{166,88,53}
\definecolor{nightblue}{RGB}{9,49,105}
\definecolor{keywords}{RGB}{207,33,46}
\definecolor{lightpurple}{RGB}{130,81,223}

\definecolor{skyblue}{RGB}{86,168,245}
\definecolor{deepblue}{rgb}{0,0,0.5}
\definecolor{deepred}{rgb}{0.6,0,0}
\definecolor{deepgreen}{rgb}{0,0.5,0}

\lstdefinestyle{mystyle}{
    backgroundcolor=\color{backcolour},
    commentstyle=\color{codegreen},
    keywordstyle=\color{keywords},
    stringstyle=\color{nightblue},
    basicstyle=\fontsize{7}{8}\ttfamily,
    breakatwhitespace=true,         
    breaklines=true,                 
    captionpos=b,                    
    keepspaces=true,                 
    numberstyle=\tiny\color{codegray},
    numbersep=2pt,                  
    showspaces=false,                
    showstringspaces=false,
    showtabs=false,                  
    tabsize=2,
    emph={model,bundle,@trace,trace,backward, instruction, code, documentation, variables, constraints,inputs,others,outputs,feedback,actual_problem_instance,variable1_name,variable2_name,variable1_value1,variable1_value2,variable2_value1,variable2_value2,feedback_1,feedback_2,node,@trace_class},
    emphstyle={\color{codepurple}},
    linewidth=1\columnwidth,
    frame=tb,    
    xrightmargin=0pt,
    xleftmargin=0.23cm,
    numbers=left,
    aboveskip=0.2cm,
    belowskip=0.1cm,
}

\usepackage{amsmath,amsfonts,bm}

\def\eqref#1{equation~\ref{#1}}

\def\1{\bm{1}}

\DeclareMathAlphabet{\mathsfit}{\encodingdefault}{\sfdefault}{m}{sl}
\SetMathAlphabet{\mathsfit}{bold}{\encodingdefault}{\sfdefault}{bx}{n}

\newcommand{\stderr}[1]{{\scriptsize \textcolor{gray}{$\pm$ #1}}}

\newif\ifanonurl
\anonurlfalse
\newcommand{\anonurl}[2]{\ifanonurl\url{#1}\else\url{#2}\fi}

\usepackage[capitalise]{cleveref}

\definecolor{workflowred}{HTML}{FFCEA6}
\definecolor{paramblue}{HTML}{ABD1FF}
\definecolor{paramblueEdge}{HTML}{68ADFF}
\definecolor{paramyellow}{HTML}{FFE5B0}
\definecolor{paramyellowEdge}{HTML}{FFD173}
\definecolor{optgreen}{HTML}{C2EBCE}
\definecolor{agentgraphorange}{HTML}{FFCEA7}
\definecolor{feedbackpurple}{HTML}{DBC9FF}
\definecolor{feedbackpurpleEdge}{HTML}{AF87FF}

\usepackage{tikz}
\usetikzlibrary{arrows.meta, positioning}

\usepackage[most,skins,theorems]{tcolorbox}

\tcbset{
  aibox/.style={
    width=\linewidth,
    top=8pt,
    bottom=4pt,
    colback=blue!6!white,
    colframe=black,
    colbacktitle=black,
    fonttitle=\small,
    fontupper=\small,
    enhanced,
    center,
    attach boxed title to top left={yshift=-0.1in,xshift=0.15in},
    boxed title style={boxrule=0pt,colframe=white,},
  }
}
\newtcolorbox{AIbox}[2][]{aibox,title=#2,#1}
\definecolor{lightblue}{rgb}{0.22,0.45,0.70}%

\tcbuselibrary{breakable}

\title{Understanding the Challenges in Iterative Generative Optimization with LLMs}

\author{%
  Allen Nie$^{1,}$\thanks{Equal contribution.\quad $^\dagger$Work done at Stanford University.\quad $^\ddagger$Work done at Microsoft Research.}\hphantom{$^{*}$}$^{,\dagger}$,
  Xavier Daull$^{2,}$\footnotemark[1]\hphantom{$^{*}$},
  Zhiyi Kuang$^{3,}$\footnotemark[1]\hphantom{$^{*}$}, \\
  \textbf{Abhinav Akkiraju}$^{4}$,
  \textbf{Anish Chaudhuri}$^{3}$,
  \textbf{Max Piasevoli}$^{5}$,
  \textbf{Ryan Rong}$^{3}$, \\
  \textbf{YuCheng Yuan}$^{3}$,
  \textbf{Prerit Choudhary}$^{3}$,
  \textbf{Shannon Xiao}$^{3}$,
  \textbf{Rasool Fakoor}$^{6}$, \\
  \textbf{Adith Swaminathan}$^{7}$,
  \textbf{Ching-An Cheng}$^{8,\ddagger}$ \\[2mm]
  $^1$Google DeepMind,
  $^2$LIS (CNRS),
  $^3$Stanford University,
  $^4$Carnegie Mellon University, \\
  $^5$Microsoft,
  $^6$AWS,
  $^7$Netflix Research,
  $^8$Google Research
}

\begin{document}

\maketitle

\begin{abstract}
Generative optimization uses large language models (LLMs) to iteratively improve artifacts (such as code, workflows or prompts) using execution feedback. 
It is a promising approach to building self-improving agents, yet in practice remains brittle: despite active research, only $9\%$ of surveyed agents used any automated optimization. 
We argue that this brittleness arises because, to set up a \emph{learning loop}, an engineer must make ``hidden'' design choices: What can the optimizer edit and what is the ``right'' learning evidence to provide at each update?
We investigate three factors that affect most applications: the starting artifact, the credit horizon for execution traces, and batching trials and errors into learning evidence.
Through case studies in MLAgentBench, Atari, and BigBench Extra Hard, we find that these design decisions can determine whether generative optimization succeeds, yet they are rarely made explicit in prior work. 
Different starting artifacts determine which solutions are reachable in MLAgentBench, truncated traces can still improve Atari agents, and larger minibatches do not monotonically improve generalization on BBEH. 
We then study these factors jointly in harness optimization on TBLite, evaluated on Terminal-Bench 2.0. We also propose a batch-size and credit-horizon rule that, calibrated on TBLite, transfers to DeepSWE after adjusting to its token costs and comes within 1\% of the best configuration tested at each context budget. 
\end{abstract}

\section{Introduction} \label{sec:intro}

Rapid advances in the capabilities of large language models (LLMs) have enabled a proliferation of software systems with the ability to perceive, plan, and reflect~\citep{measuring-ai-ability-to-complete-long-tasks}. 
Recent work has shown that LLMs have the ability to generate and revise program workflows to optimize an objective~\citep{yang2024large}, such as increasing compute throughput or decreasing latency of hardware accelerator kernels~\citep{ouyang2025kernelbench,lange2025towards,wei2025astra,zhang2025accelopt}, design novel algorithms for search and matrix multiplication~\citep{wei2024improving,press2025algotune,novikov2025alphaevolve}, exploiting security vulnerabilities~\citep{zhang2024cybench,chaudhuri2025optimizing}, and propose therapeutic candidates for diseases~\citep{ghareeb2025robin}. This ability to optimize objectives, combined with LLMs continuing to approach human-level performance in producing complex programs~\citep{jimenez2023swe,wang2024openhands,el2025competitive,wei2025swe}, gives rise to an emerging class of software that automatically changes its own behavior to achieve a desired outcome. 

Automated generation and optimization with LLMs have been adopted broadly in two types of applications.
The first is to use the LLM to repeatedly modify a software program to improve on a metric, such as writing kernels that have low compute latency~\citep{baronio2025kevin,wei2025astra,zhang2025accelopt}, creating automated ML pipelines to train models that achieve high test accuracy~\citep{huang2023mlagentbench,chan2024mle,toledo2025ai}, writing a script that can exploit security vulnerabilities~\citep{zhang2024cybench,wang2026exploitgym}, and optimizing agent harnesses~\citep{lee2026metaharness}. The second type uses LLMs to modify another LLM system to achieve desired behavior, also using metrics like success rate, through prompt tuning and direct code revisions~\citep{khattab2023dspy,yuksekgonul2024textgrad,cheng2024trace,wang2024correctly,zhang2025recursive}. The underlying mechanism in both types of applications is the same: construction of an LLM-based \textit{generative optimization} process to ingest feedback and modify an existing system, which we describe as a \textit{learning loop} in Section~\ref{sec:core_concepts}. 

However, despite the popularity in research, LLM-based generative optimization has not been widely adopted
in production. \citet{pan2025measuring} report that the current development of agentic systems remains largely human-driven, with only 9\% of surveyed systems employing any form of automated design, including simple LLM-assisted prompt tuning. 
By contrast, using LLMs to optimize programs has been more successful in specialized domains~\citep{toledo2025ai,novikov2025alphaevolve}. 
In a field where end-to-end automation that scales with compute is a highly sought-after objective~\citep{sutton2019bitter,hendrycks2025definition}, the lack of wider adoption
is puzzling, pointing to a potential gap between the high ideal of an automated LLM optimizer and reality, especially for optimizing LLM agentic systems.

This lack of adoption is not a consequence of inadequate infrastructure support or insufficient software abstraction. On the contrary, over the past two years, a rich ecosystem of powerful libraries for building agentic systems has emerged~\citep{khattab2023dspy,wu2023autogen,langgraph2024,cheng2024trace,li2025flow}. Many of these libraries offer mechanisms for 
automatically optimizing different targets,
ranging from tuning prompts~\citep{khattab2023dspy,yuksekgonul2024textgrad} to program synthesis~\citep{cheng2024trace}. Learning loop has become an accepted and more ubiquitous design in OpenClaw~\citep{steinberger2026openclaw}, Claude Code~\citep{anthropic2025claudecode}, and Codex~\citep{openai2025codex}. Most of them have received considerable attention from agent engineers, suggesting that the low adoption rate cannot be attributed solely to limited awareness or inadequate software.

In this paper, we hypothesize that the low adoption stems from the hidden difficulty of setting up the \textit{learning loop}. Our experiments show that getting the learning loop right requires substantial engineering effort and/or guesswork. This setup burden is a major hurdle for productionization, which requires simple, universal solutions across application domains.
In Section~\ref{sec:core_concepts}, we first introduce two core concepts that affect the learning loop: system initialization and learning context construction.
Then we examine three case studies to isolate and highlight the challenges related to these core concepts, and show how different design decisions can impact the final performance. In Section~\ref{sec:interactive}, we show how different system initializations affect the final model quality of an ML training pipeline.
In Section~\ref{sec:atari}, using the example of an Atari game-playing program, we show that engineers can specify a credit horizon that is shorter than the full gameplay trajectory and still obtain programs that achieve high reward on a full playthrough, but also that updating the system too frequently with too short a credit horizon leads to worse performance. In Section~\ref{sec:batch}, we show that the number of examples placed in the learning context matters for optimizing the prompt of an LLM call.
Finally, in Section~\ref{sec:harness-experiments}, we jointly vary the design choices from all three case studies while optimizing an agent harness on terminal tasks (TBLite), and find that performance is sensitive to each of them. We also show that a simple rule for splitting a fixed context budget between batch size and credit horizon can transfer across domains: calibrated on TBLite and evaluated on software engineering tasks (DeepSWE), it picks a configuration close to the best one tested at each budget.

Interestingly, many of the challenges in setting up a learning loop for LLM-based optimization parallel well-studied concepts in machine learning. The \textbf{starting artifact problem} resembles neural network architecture~\citep{zoph2016neural} and weights initialization~\citep{glorot2010understanding}, where different starting points determine which solutions are reachable. The \textbf{credit horizon problem} mirrors horizon debates in episodic reinforcement learning~\citep{arjona2019rudder,cheng2021heuristic,laidlaw2023bridging}
and truncated back-propagation through time~\citep{tallec2017unbiasing,shaban2019truncated}. The \textbf{experience batching problem} parallels batch size selection in stochastic gradient descent, where the number of examples aggregated per update affects both learning dynamics and generalization~\citep{smith2017don}. However, unlike traditional ML where practitioners have developed theoretical guidance and/or heuristics, the learning loop design space for generative optimization remains largely unexplored. 
We suggest that the challenges of LLM-based generative optimization are similar to the challenges in traditional machine learning and can be studied systematically rather than treated as ad hoc engineering. Our investigation also gives rise to the need of constructing task-specific, customized learning context for optimization problems, as seen in recent works~\citep{zhang2026stateless,tan2026scaling}.

\section{Building a Learning Loop}
\label{sec:core_concepts}

We start by describing the concept of a \textit{learning loop}\footnote{We give a more rigorous description in Appendix~\ref{sec:algorithm}, connecting to the framework by \citet{cheng2024trace}.}, which is ubiquitous in a wide range of LLM-based generative optimization applications
 (Figure~\ref{fig:learning_loop}). We are given an initial system ({\color{workflowred}\rule{1em}{0.6em}}) that takes an input and produces an output, and an oracle to give feedback ({\color{feedbackpurple}\rule{1em}{0.6em}}) that can serve as a signal for optimizing. 
In theory, these two terms define a learning problem conceptually. However, in practice, more details are needed to implement an actual learning loop with an LLM optimizer: 
 \begin{wrapfigure}{o}{0.25\linewidth}
    \hspace{-5mm}
    \vspace{-3mm}
    \centering
    \includegraphics[width=\linewidth]{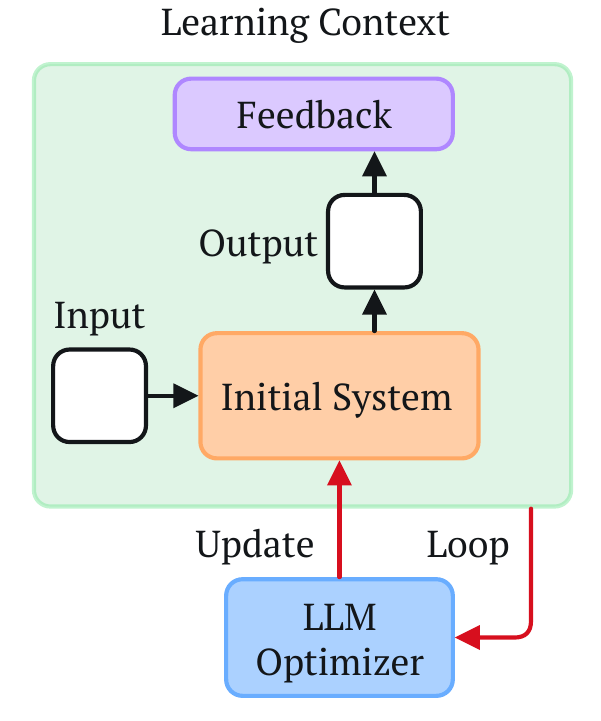}
    \vspace{1mm}
    \caption{The learning loop of generative optimization.}
    \label{fig:learning_loop}
    \vspace{-7mm}
\end{wrapfigure}
What exactly should be included in a \textit{learning context} (i.e. the message to send in the LLM's API call) so that the LLM optimizer can make effective updates?
As shown in Figure~\ref{fig:learning_loop}, a typical learning context includes input, output, feedback, and initial/current system.
In addition, other common information includes task background and the optimizer's past experiences of successes/failures.
Once designed, the content of the learning context will be dynamically updated during optimization to reflect the up-to-date optimization status.

Similar to how a numerical optimization process depends on its initial condition and optimization step function, 
we can also break down the analysis of the learning loop into: What is the starting point? What information should be provided to the LLM optimizer at each step?%

\begin{figure*}[t]
    \centering
    \includegraphics[width=0.85\linewidth]{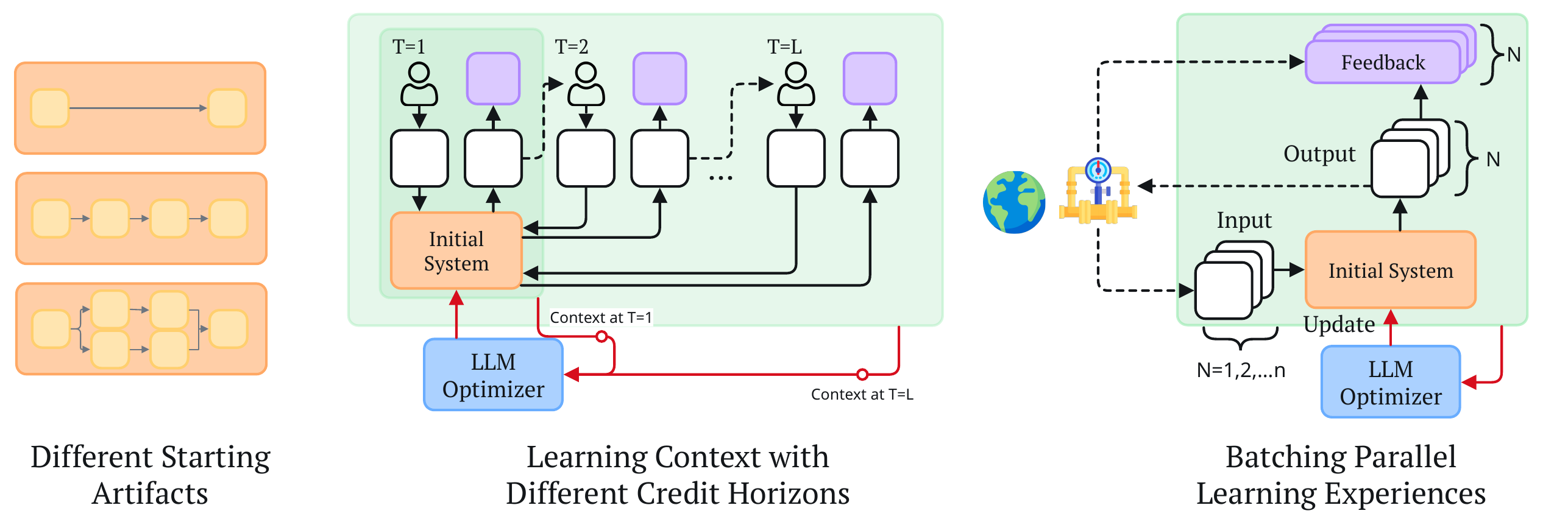}
    \caption{\textbf{Three Key Decisions for Implementing a Learning Loop}. To set up iterative generative optimization, an agent engineer must make three core decisions. For the initial system: What (1) \textbf{starting artifacts} ({\color{paramyellow}\rule{1em}{0.6em}} instructions,  files, specs) to provide? Different initializations can lead to different solution spaces. For the learning context ({\color{optgreen}\rule{1em}{0.6em}}): What learning evidence to provide to the LLM optimizer -- (2) how many steps to include per trial (\textbf{credit horizon}) and (3) how many trials to batch together (\textbf{experience batching}).}
    \label{fig:optimization-three-problems}
\end{figure*}

\vspace{-2mm}
\paragraph{Initial System (Starting Artifact)} This includes the initial code, prompt, and files that make up the system. The initial system can also consist of documentation and design sketches that are sent to an LLM programmer (such as GPT Codex, Claude Code, or Gemini CLI). The initial system is a choice made by the engineer. In addition, the engineer must also determine which part of the system can be changed by the LLM optimizer and which part is constrained, e.g., the whole codebase or just certain functions.
In Section~\ref{sec:interactive}, we show that different designs lead to major differences in the quality of ML pipelines generated by an LLM optimizer, giving rise to the \textbf{starting artifact problem}.

\vspace{-2mm}
\paragraph{Learning Context (Credit Horizon)}
Prior work often \textit{assumes} either that the context is sufficient for the LLM optimizer to improve successfully~\citep{cheng2024trace} or that the feedback \textit{usually} contains useful signal~\citep{nie2024importance,xu2025provably}. 
However, what context is necessary or sufficient can be ambiguous in practice, especially for multi-step problems, where the system being optimized is used multiple times sequentially.
The agentic system community has already started to explore solutions for this problem, with early work from \citet{zhang2025agentic,sun2025scaling,ye2025agentfold,zhang2025recursive}. 
More generally, it is unclear how many steps of the process's execution trace should be included in the learning context for the optimizer. Should we optimize the agent from instantaneous feedback, or should we only optimize it after all feedback in the multi-step process has been observed? 
In Section~\ref{sec:atari}, we study an exemplar version of this problem by optimizing an Atari game-playing program, where both dense short-term (after each action) and long-term reward (after the game ends) can be used to modify the agentic system's behavior. We see that for four out of eight games, optimizing for short-term dense reward produced systems that also performed comparably well for the full episode. We call this the \textbf{credit horizon problem}, which is related to the effective horizon in RL \citep{laidlaw2023bridging,cheng2021heuristic}.

\vspace{-2mm}
\paragraph{Learning Context (Experience Batching)} 
After deciding the credit horizon (the number of steps to include for the optimizer), another key decision is about how to present the experience of successes and failures of \textit{independent} trials to the optimizer. The community has previously used words like memory to study this phenomenon~\citep{memento,chhikara2025mem0buildingproductionreadyai,ouyang2025reasoningbank,zhang2025accelopt}. However, most of these works focus on techniques for ``retrieving'' relevant memories. We instead focus on an even simpler but more fundamental aspect: the amount of in-context experience across independent trials provided to the optimizer through the mechanism of ``batching.''
In Section~\ref{sec:batch}, we look at this problem in a setup inspired by batched stochastic gradient descent. We study a task involving 
optimization of a prompted LLM system. We explore different numbers of  (input, output, feedback) triplets to put in the learning context. We observe that ``batch size'' affects whether the LLM optimizer can find a prompt that does well on a hidden test set. We show that the optimal number of triplets is different for each task, giving rise to the \textbf{experience batching problem}. 

In our experiments, we find that the best configurations to address all three problems are task-dependent. Different tasks require different setups in order to achieve the best results. 
All three problems introduce complexities and require the agent engineer to make nuanced decisions.

We note that the learning loop can be implemented with any LLM optimization and search algorithms~\citep{novikov2025alphaevolve,pan2025learning,lange2025shinkaevolve,agrawal2025gepa,ren2026polcastochasticgenerativeoptimization} and the main focus of our paper is to study the factors that impact the learning loop, not the small differences between individual libraries. 
Our goal is to investigate other lesser-known factors that critically impact the success of the optimization in order to provide practical guidelines beyond a mere choice of a ``search'' algorithm.
\vspace{-1mm}
\section{Related Work}
\label{sec:related}
\vspace{-1mm}

\paragraph{Learning Loop in Self-Improving Agents} The concept of a \textit{loop} is widely discussed in the LLM agent community, commonly referred to as an agent loop~\citep{zhao2025agent,bolin2026unrollingcodexagentloop}, a sampling loop~\citep{anthropic_claude_quickstarts_loop_py}, or a ``Ralph'' loop~\citep{huntley2026loop}. These loops typically enable abilities like self-debugging~\citep{chen2023teaching}, self-correction~\citep{xiong2025self}, and self-refinement~\citep{madaan2023self} to make agents more likely to succeed within a single task execution. 
In contrast, rather than optimizing for the highest success rate on an individual task, our \textit{learning loop} is designed for continual learning through repeated trial and error~\citep{huang2024self,monea2024llms}. Compared with a within-task agent loop, our \textit{learning loop} accumulates experience across tasks, where the success or failure of any single attempt is secondary to the agent's eventual mastery.

\paragraph{Context Engineering} Building the right context for LLMs has received significant attention in recent works. Beyond compression of overly long inputs, carefully constructed context can substantially improve performance across diverse tasks~\citep{chen2025flora,zhang2025agentic}. Related concepts have been explored under the term ``memory''~\citep{wang2024agent,ouyang2025reasoningbank,memento}, focusing on techniques to retrieve and manage relevant past information.
We use the term \textit{learning context} to refer to the evidence provided to an LLM optimizer for system improvement. We focus specifically on two aspects that have received little systematic investigation: the horizon of multi-step traces and the number of independent traces. These two factors impact LLM self-improvement loops but have not been discussed in depth in prior work.

\paragraph{Agentic Libraries with Learning Loops} Many frameworks enable LLMs to iteratively modify systems~\citep{yang2024large}, particularly for prompt optimization~\citep{khattab2023dspy,cheng2024trace,yuksekgonul2024textgrad,wang2024correctly}. These works implement learning loops as candidate-selection procedures, using techniques like cross-validation~\citep{khattab2023dspy} or Pareto optimization~\citep{conway2025syftr,agrawal2025gepa}. However, these works primarily showcase successful applications rather than investigating the design choices and instabilities that make learning loops difficult to implement.

\section{Factors on Learning-Loop Design}
\label{sec:factors}
\label{sec:spotlights}

We first examine how each factor impacts the final solution discovered by a learning loop. 
We implement the ablations in this section with Trace and OptoPrime~\citep{cheng2024trace} for ease of experimentation, while the components of a learning loop are ubiquitous across different frameworks and implementations. We first investigate the three factors separately across a wide range of tasks, from machine learning engineering, to writing programs that play Atari games, to prompt optimization. The comparisons below highlight the resulting differences, with implementation and evaluation details in the corresponding appendices.

\subsection{Starting Artifact in MLAgentBench}
\label{sec:interactive}

The starting artifact defines the program structure that the optimizer can revise. On Housing Price and Spaceship Titanic from MLAgentBench~\citep{huang2023mlagentbench}, we compare a single \texttt{train\_model} function with a pipeline of separate functions. Both initializations provide equivalent docstring information and use the same outer optimization protocol. This lets us examine the effect of exposing the ML pipeline as one editable block or as several coordinated components.

\begin{figure}[htbp]
    \centering
    \begin{minipage}[c]{0.53\linewidth}
        \centering
        \begin{subfigure}[h]{0.48\linewidth}
            \centering
            \includegraphics[width=0.95\linewidth]{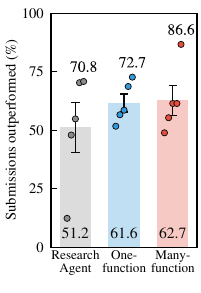}
            \captionsetup{margin={23.05pt,0pt}}
            \subcaption{Spaceship Titanic}
            \label{fig:spaceship_one_v_multi}
        \end{subfigure}%
        \hfill
        \begin{subfigure}[h]{0.48\linewidth}
            \centering
            \includegraphics[width=0.95\linewidth]{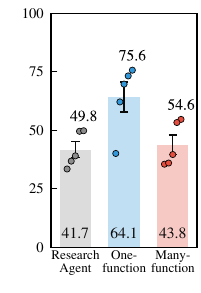}
            \captionsetup{margin={23.05pt,0pt}}
            \subcaption{Housing Price}
            \label{fig:housing_one_v_multi}
        \end{subfigure}
    \end{minipage}%
    \hfill
    \begin{minipage}[c]{0.45\linewidth}
        \centering
        \includegraphics[width=\linewidth]{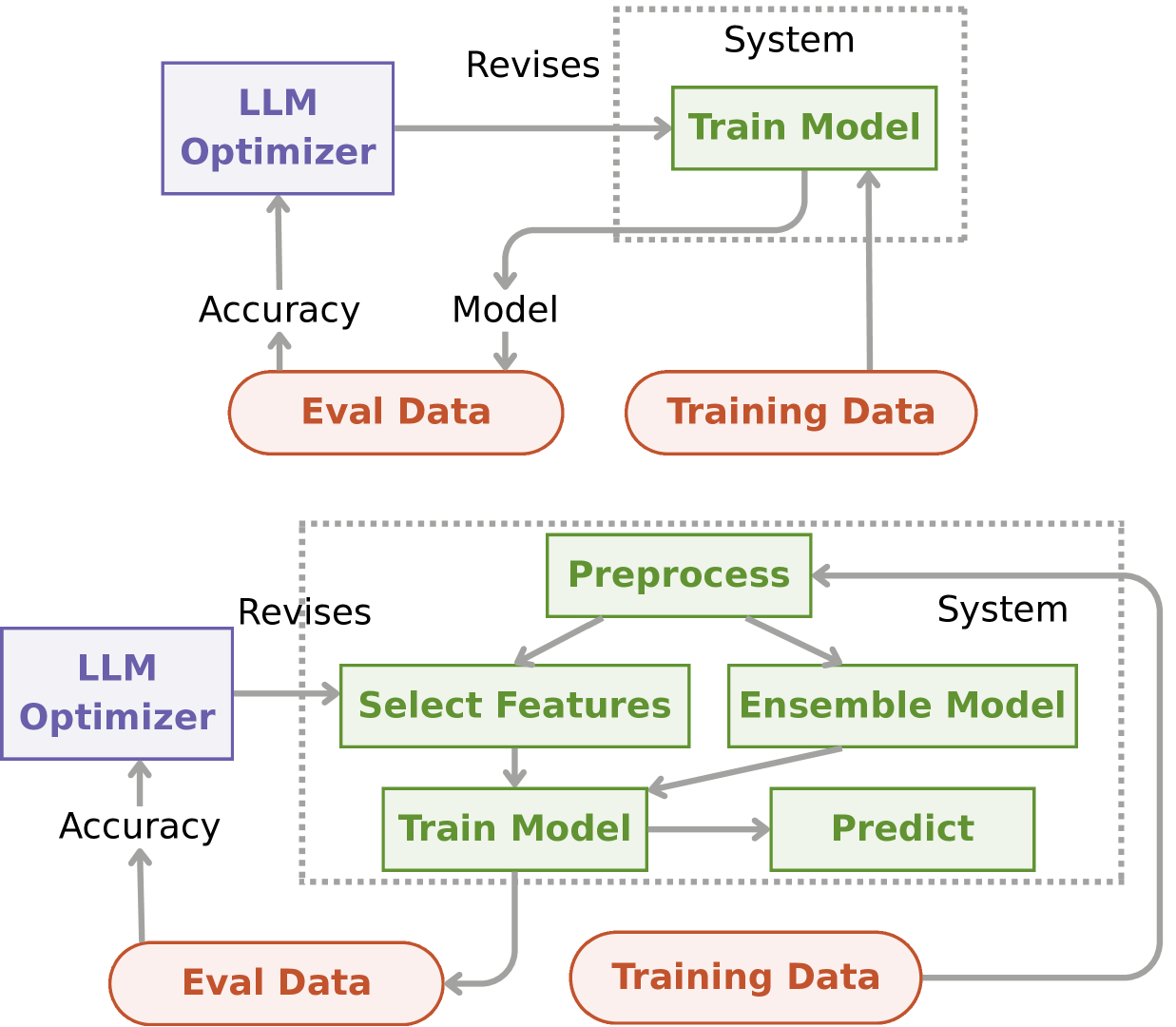}
    \end{minipage}
    \caption{\textbf{Starting artifact in MLAgentBench.} Left: share of Kaggle leaderboard submissions each run outperforms. Dots are the five runs, bars their mean with $\pm$1 standard error, and the number above each arm its best run; ResearchAgent~\citep{huang2023mlagentbench} is the baseline. Right: the one-function and many-function workflow designs, which provide equivalent docstring information.}
    \label{fig:mlagent_leaderboard_result}
    \label{fig:ml_agent_one_vs_many_funcs}
\end{figure}

Figure~\ref{fig:mlagent_leaderboard_result} shows that the ordering of these starting artifacts reverses across the two tasks. Over five trials, the best many-function pipeline on Spaceship Titanic surpasses 86.6\% of leaderboard submissions, compared with 72.7\% for the one-function pipeline. On Housing Price, the one-function pipeline surpasses 75.6\% of submissions, compared with 54.6\% for the many-function pipeline.

The figure also separates average performance from the best discovered pipeline. Table~\ref{tab:mlagent_main_result} in Appendix~\ref{sec:app:mlagent} reports the underlying submission scores. On Spaceship Titanic, the mean submission accuracies are close (79.65\% versus 79.69\%), even though the best runs occupy different leaderboard percentiles. The starting artifact therefore affects what the optimizer discovers in these runs, with the size of the difference depending on the task and summary metric. The appendix also gives the feedback and evaluation protocol.

\subsection{Credit Horizon in Atari}
\label{sec:atari}

The next choice concerns how much of a sequential execution trace the optimizer should see before revising the system. We study Python policies that act on structured OCAtari states~\citep{delfosse2024ocatari}. From the same starting policy, one condition updates after a single observation, action, and immediate reward. The other collects a longer rollout before each update. Both learned policies are evaluated over longer gameplay trajectories.

\begin{figure}[htbp]
    \centering
    \includegraphics[width=\linewidth]{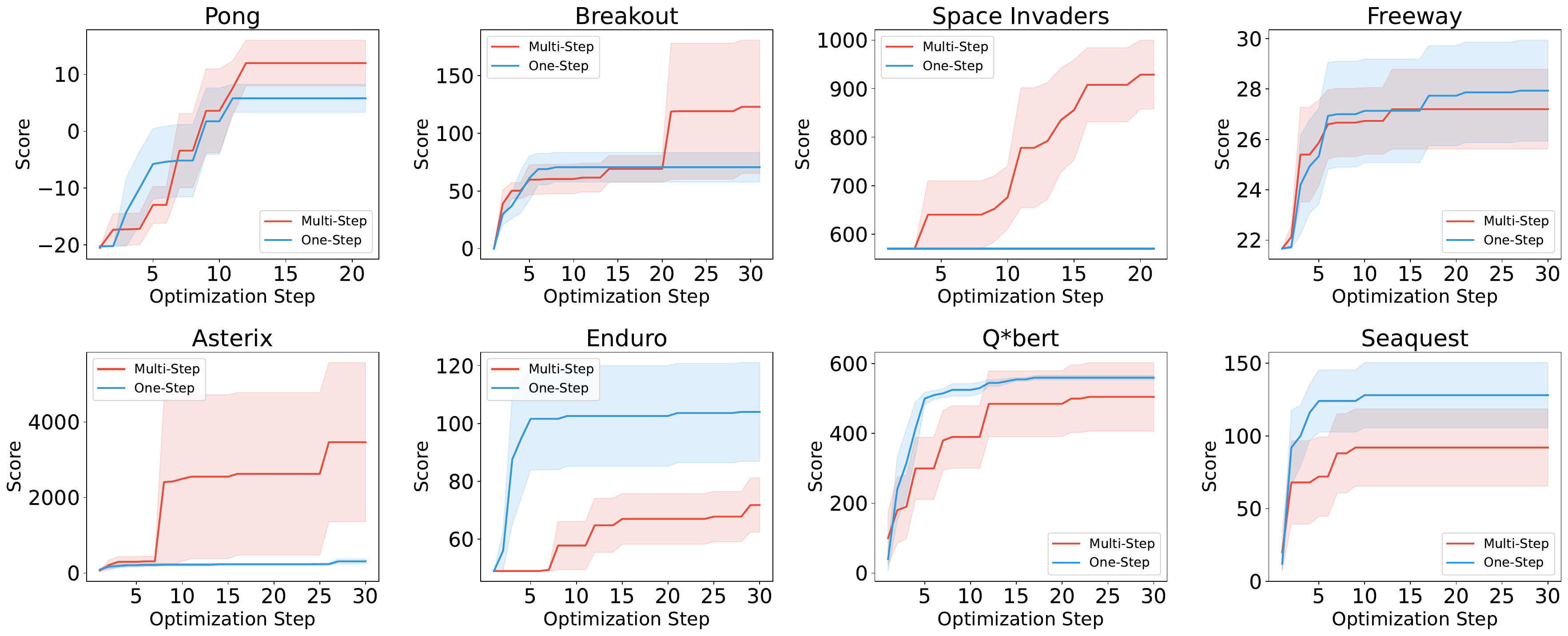}
    \caption{\textbf{Credit horizon in Atari.} One-step and multi-step optimization across five trials per game, using the same starting artifact. Longer rollouts perform better in four of the eight games.}
    \label{fig:atari-learning-curve}
\end{figure}

\noindent
\begin{minipage}[t]{0.48\linewidth}
\vspace{0pt}
\setlength{\parskip}{3pt}
The learning curves in Figure~\ref{fig:atari-learning-curve} show that neither update configuration dominates. Across five trials per game, multi-step optimization performs better in Pong, Breakout, Space Invaders, and Asterix, while one-step optimization performs better in Freeway, Enduro, Q*bert, and Seaquest.

A longer trace can expose the consequences of earlier actions, which may help in Space Invaders, where effective play requires coordinating shooting and movement under delayed consequences. Games such as Freeway can instead benefit from more frequent updates based on short-horizon feedback. Appendix~\ref{sec:app:atari} gives the rollout lengths, feedback, evaluation configurations, and additional analysis.

The same runs also let us compare the training efficiency of generative optimization with object-centric deep RL baselines (Figure~\ref{fig:atari-training-efficiency}). In these experiments, the code policies reach their best observed performance in substantially less wall-clock time. Attained scores vary across games, so the figure presents runtime and normalized reward together (Appendix~\ref{sec:app:deep-rl-result}).
\end{minipage}\hfill
\begin{minipage}[t]{0.5\linewidth}
\vspace{0pt}
\captionsetup{type=figure}
\centering
    \includegraphics[width=\linewidth]{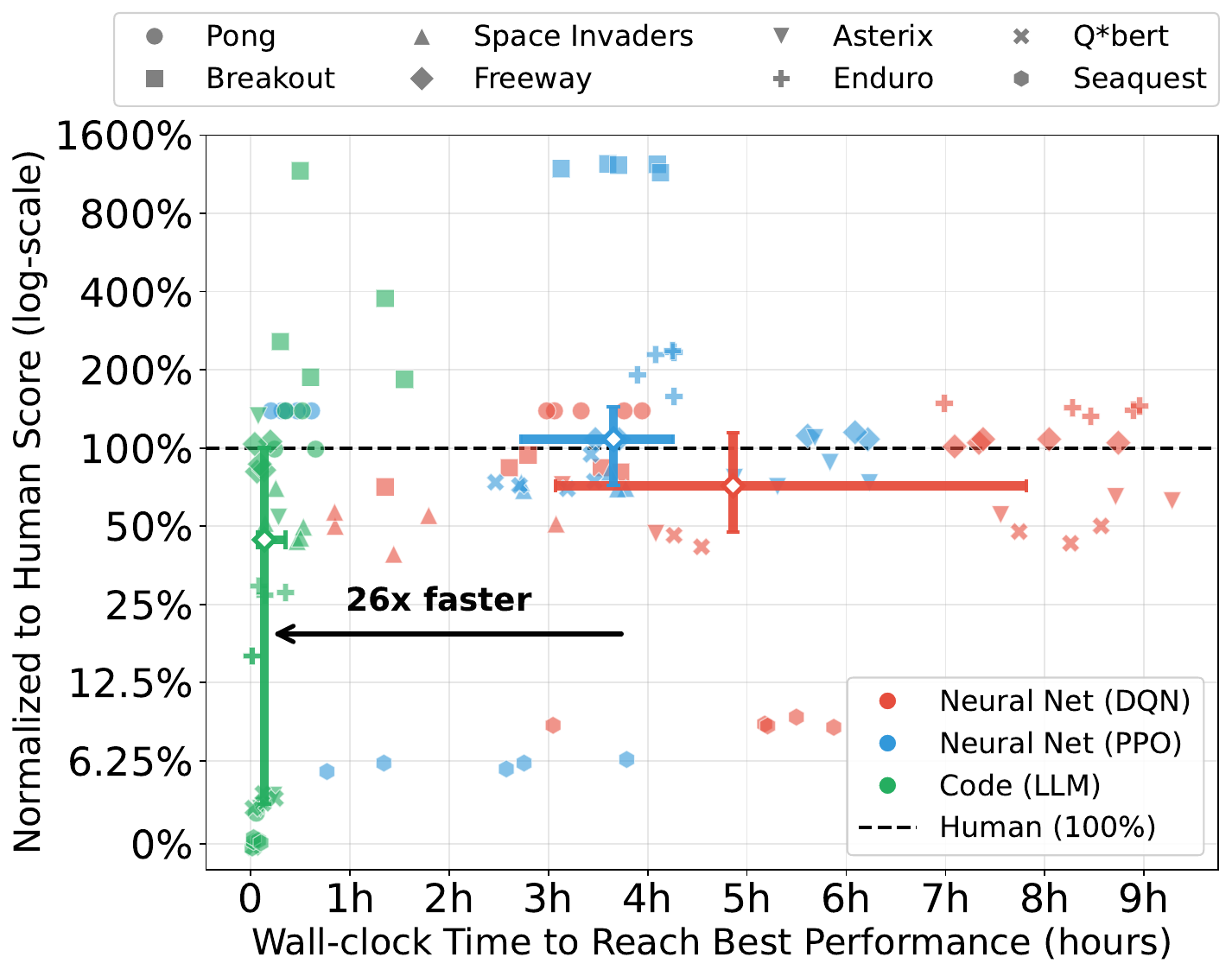}
    \caption{\textbf{Training Efficiency Comparison}. Deep RL training builds on CleanRL~\citep{huang2022cleanrl}; PPO uses ten parallel environments for sampling while DQN and LLM use a single environment. Both PPO and DQN use batch training on a single A100 GPU. All algorithms use object-centric inputs from OCAtari. The deep RL algorithms convert this information into numeric vectors, whereas the LLM takes the raw dictionary without additional annotation. Scores are normalized to map 0\% to random policy on the task and 100\% to human.}
    \label{fig:atari-training-efficiency}
\end{minipage}
\par\medskip

\subsection{Experience Batching in BigBench Extra Hard}
\label{sec:batch}

The engineer must also decide how much experience to aggregate across independent examples. On BigBench Extra Hard (BBEH)~\citep{kazemi2025big}, we optimize a prompt and answer-extraction function, concatenating the traces from 1, 3, or 5 examples into each update. All configurations receive the same number of optimizer updates. The learned prompt and extraction code must then generalize to held-out questions.

\begin{figure}[htbp]
    \centering
    \includegraphics[width=\linewidth]{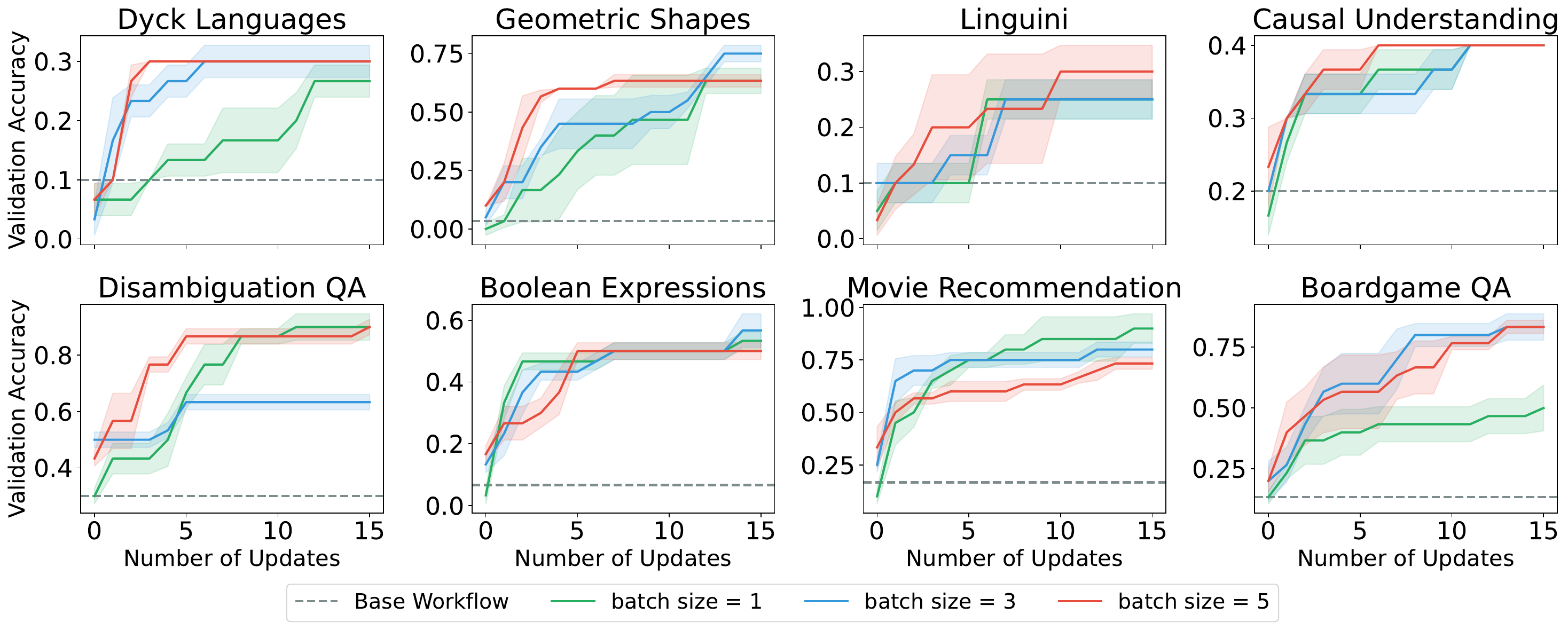}
    \caption{\textbf{Experience batching in BBEH.} Validation accuracy across optimizer updates for batch sizes 1, 3, and 5. Curves aggregate three trials, with shaded standard errors. Larger batches often improve early learning but can plateau.}
    \label{fig:minibatch-learning-curve-full}
\end{figure}

Figure~\ref{fig:minibatch-learning-curve-full} shows how batch size changes the learning process. Larger batches often improve validation accuracy early, but can plateau, while smaller batches produce noisier trajectories that sometimes continue improving. We therefore examine held-out accuracy alongside these learning curves.

\begin{table}[htbp]
\centering
\small
\caption{
\textbf{Held-out accuracy across batch sizes.} Highest mean accuracy among the three batch sizes is bolded per task. Gray values give standard errors over three trials. The backend is Claude Sonnet-3.5-v2, with typically 175+ test examples per task.
}
\vspace{1mm}
\label{tab:structure_lang_tasks}
\begin{tabular}{@{}lccc@{}}
\toprule
& \multicolumn{3}{c}{Batch size} \\
\cmidrule(l){2-4}
Task & 1 & 3 & 5 \\
\midrule
Dyck Languages       & 0.183 \stderr{0.049} & 0.063 \stderr{0.010} & \textbf{0.190} \stderr{0.031} \\
Geometric Shapes     & 0.343 \stderr{0.039} & \textbf{0.389} \stderr{0.040} & 0.200 \stderr{0.099} \\
Linguini             & 0.149 \stderr{0.024} & \textbf{0.234} \stderr{0.012} & 0.170 \stderr{0.030} \\
Causal Understanding & 0.375 \stderr{0.146} & 0.408 \stderr{0.097} & \textbf{0.531} \stderr{0.018} \\
Disambiguation QA    & \textbf{0.537} \stderr{0.036} & 0.295 \stderr{0.091} & 0.526 \stderr{0.035} \\
Boolean Expressions  & 0.177 \stderr{0.005} & \textbf{0.238} \stderr{0.006} & 0.154 \stderr{0.034} \\
Movie Recommendation & \textbf{0.889} \stderr{0.038} & 0.683 \stderr{0.119} & 0.810 \stderr{0.016} \\
Boardgame QA         & \textbf{0.341} \stderr{0.032} & 0.278 \stderr{0.009} & 0.276 \stderr{0.007} \\
\bottomrule
\end{tabular}
\end{table}

Table~\ref{tab:structure_lang_tasks} shows that the preferred batch size differs across tasks. Size 1 gives the highest test accuracy on Movie Recommendation, size 3 on Geometric Shapes, and size 5 on Causal Understanding. Thus, the broader coverage supplied by a larger batch does not consistently yield a better final prompt in this experiment. Together with the Atari comparison, this raises the question of how to allocate learning context between independent examples and steps within each example. Appendix~\ref{sec:app:mini-batch-learning} gives the data splits, training protocol, and comparisons with DSPy and LangGraph to show that batch size also impacts the optimization result of those two frameworks.

\begin{AIbox}{Summary}

Modularizing the \textbf{starting artifact} helped some MLE tasks but hurt others, and even short, incomplete \textbf{credit horizons} produced successful policies.
For \textbf{experience batching}, larger batches can speed early learning without reliably improving held-out accuracy.
\end{AIbox}

\section{A Unified Experiment in Harness Optimization}
\label{sec:harness-experiments}

Section~\ref{sec:factors} highlights each factor in a separate setting. We now study the three factors jointly in harness optimization on TBLite, with evaluation on Terminal-Bench~2.0. We then transfer the starting-artifact choice and guidance for allocating learning context from TBLite to DeepSWE, testing whether they produce harnesses competitive with the best configurations tested at each context budget.

\subsection{Sensitivity to the Three Factors}
We rewrite agent source code with a direct Claude Sonnet~5 call, and the agent executes tasks with Claude Haiku~4.5. We use 35 TBLite~\citep{OpenThoughts-TBLite} tasks for training, 15 for validation and harness selection, and reserve 50 for testing. Terminal-Bench~2.0~\citep{tbench_2025} is held out for transfer. The sweep scores use fixed subsets of 15 test tasks per benchmark. Appendix~\ref{sec:app:harness} gives the evaluation protocol and a comparison with other Haiku~4.5 harnesses.

\paragraph{Search grid} We cross two starting artifacts---a single-file harness and a four-module harness---with batch sizes of 1, 5, or all 35 training tasks and horizons of 10\%, 50\%, or 100\% of each trajectory. All files are editable in both cases; Appendix~\ref{sec:app:harness:sweep} gives the module structure and context-truncation details.
We additionally test Mini-SWE-Agent and Terminus-2 as starting artifacts at batch size 5 and a 50\% horizon.

\begin{table}[htbp]
\centering
\small
\setlength{\tabcolsep}{3.5pt}
\begin{minipage}[t]{0.47\linewidth}
\centering
\caption{Scores across the terminal harness sweep. The highest reported score on each benchmark is bolded.}
\label{tab:harness-grid}
\begin{tabular}{@{}ccrrrr@{}}
\toprule
& & \multicolumn{2}{c}{Single-file} & \multicolumn{2}{c}{Modular} \\
\cmidrule(lr){3-4}\cmidrule(lr){5-6}
Batch & Horizon & TBLite & TB2 & TBLite & TB2 \\
\midrule
1   & 10\%  & 0.533 & 0.200 & 0.400 & 0.133 \\
1   & 50\%  & 0.467 & 0.333 & 0.600 & 0.267 \\
1   & 100\% & 0.467 & 0.133 & 0.333 & 0.133 \\ \midrule
5   & 10\%  & 0.467 & 0.133 & 0.533 & 0.333 \\
5   & 50\%  & \textbf{0.800} & 0.267 & 0.600 & 0.200 \\
5   & 100\% & 0.400 & 0.133 & 0.600 & \textbf{0.400} \\ \midrule
35 & 10\%  & 0.600 & 0.333 & 0.667 & 0.200 \\
35 & 50\%  & 0.533 & 0.267 & 0.600 & 0.200 \\
35 & 100\% & 0.467 & 0.333 & 0.533 & 0.200 \\
\bottomrule
\end{tabular}
\end{minipage}
\hfill
\begin{minipage}[t]{0.495\linewidth}
\centering
\caption{One-factor sensitivity around the single-file, batch-5, 50\%-horizon setting. Each row varies one factor and fixes the other two. Ranges span the tested settings; $\Delta$ is the difference between the highest and lowest scores. We conclude that the harness optimization process is most sensitive to the starting artifact (initial harness). Appendix~\ref{sec:app:harness:sweep} describes the starting artifacts.}
\label{tab:harness-factor-sensitivity}
\begin{tabular}{@{}lrrrr@{}}
\toprule
& \multicolumn{2}{c}{TBLite} & \multicolumn{2}{c}{TB2} \\
\cmidrule(lr){2-3}\cmidrule(lr){4-5}
Factor & Range & $\Delta$ & Range & $\Delta$ \\
\midrule
Artifact & 0.333--0.800 & \textbf{0.467} & 0.267--0.400 & \textbf{0.133} \\
Horizon & 0.400--0.800 & 0.400 & 0.133--0.267 & \textbf{0.133} \\
Batch size & 0.467--0.800 & 0.333 & 0.267--0.333 & 0.067 \\
\bottomrule
\end{tabular}
\end{minipage}
\end{table}

\paragraph{Sensitivity and interactions} At the reference setting, varying any of the three factors changes the observed TBLite score by at least 0.333 (Table~\ref{tab:harness-factor-sensitivity}). The effects also depend on their combination (Table~\ref{tab:harness-grid}). In this search grid, the highest TBLite score, 0.800, comes from a single-file harness with batch size 5 and a 50\% horizon. The highest Terminal-Bench~2.0 score, 0.400, comes from a modular harness with the same batch size and a full horizon. At batch size 1, extending the horizon from 10\% to 50\% reduces the single-file harness's TBLite score from 0.533 to 0.467 but increases the modular harness's score from 0.400 to 0.600. Thus, even the direction of the credit horizon's effect depends on another factor, the starting artifact. This motivates us to come up with some practical guidance so that an optimizer can balance these choices effectively. 

\subsection{Transferring Learning-Loop Choices Across Domains}
\label{sec:harness-budget}

We have shown that on a single domain, the performance of the final optimized harness is sensitive to all three factors, albeit to different degrees. However, this alone only gives us a per-domain search grid that is very expensive to search through. We now ask whether the choice of learning context factors identified in one domain transfers to another domain---in our case, from TBLite / Terminal-Bench~2.0 to DeepSWE~\citep{huang2026deepswe}.

We first highlight the constraint of a fixed context budget $C$. Both batch size and credit horizon share this budget: including more tasks leaves less room for the trace of each task. If the optimizer sees $B$ tasks and up to $L$ steps per task, we approximate the number of tokens in this context as
\[
    C \approx c_{\mathrm{code}} + B(c_{\mathrm{task}} + c_{\mathrm{step}}L),
\]
where $c_{\mathrm{code}}$ is the artifact size, $c_{\mathrm{task}}$ is the fixed overhead per task such as task instruction, and $c_{\mathrm{step}}$ is the cost of one logged step. We estimate these quantities from roughly ten pilot rollouts of the initial harness. We define a new statistic $\rho=c_{\mathrm{task}}/c_{\mathrm{step}}$, which indicates a balance between the diversity of tasks presented to the LLM optimizer (governed by batch size) and details of logged agent behavior on the task (governed by credit horizon).

\begin{figure}[!htb]
\centering
\includegraphics[width=0.49\linewidth]{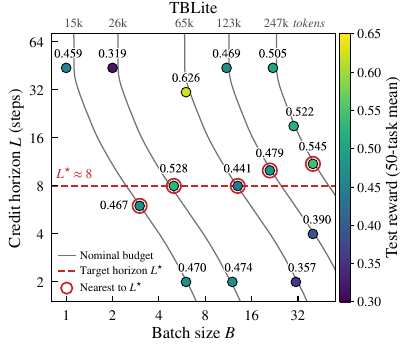}
\hfill
\includegraphics[width=0.49\linewidth]{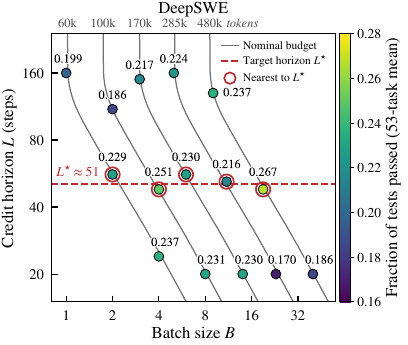}
\caption{Batch size and credit horizon under fixed optimizer-context budgets. Points show tested configurations and their scores; gray curves indicate approximate equal context budget levels. Dashed red lines mark the target horizon $L^\star$ defined below, and red circles mark the tested configuration nearest it at each budget.}
\label{fig:harness-iso-context}
\end{figure}

\paragraph{Fixed-budget comparison} We evaluate three batch--horizon configurations at each of five token budgets in each domain (Figure~\ref{fig:harness-iso-context}). We use 40 TBLite tasks for training, 10 for validation, and 50 for testing; on DeepSWE, we use 40 for training, 20 for validation, and 53 for testing. Appendix~\ref{sec:app:harness:budget} reports all 30 configurations.

\begin{table}[h]
\centering
\small
\caption{Per-budget choices for the transferred horizon rule. Each pair $(B,L)$ is the tested configuration nearest the target horizon at that budget. Best-in-Budget is the highest test score among the three configurations at the same budget.}
\vspace{1mm}
\label{tab:deep-swe-horizon-transfer}
\begin{tabular}{@{}rcccrccc@{}}
\toprule
\multicolumn{4}{c}{TBLite ($L^\star\approx8$)} & \multicolumn{4}{c}{DeepSWE ($L^\star\approx51$)} \\
\cmidrule(lr){1-4}\cmidrule(lr){5-8}
Budget & $(B,L)$ & Rule picks & Best-in-Budget & Budget & $(B,L)$ & Rule picks & Best-in-Budget \\
\midrule
15k  & $(3,6)$   & 0.467 & 0.470 & 60k  & $(2,56)$  & 0.229 & 0.237 \\
26k  & $(5,8)$   & 0.528 & 0.528 & 100k & $(4,48)$  & 0.251 & 0.251 \\
65k  & $(13,8)$  & 0.441 & 0.626 & 170k & $(6,56)$  & 0.230 & 0.230 \\
123k & $(21,10)$ & 0.479 & 0.479 & 285k & $(11,52)$ & 0.216 & 0.224 \\
247k & $(40,11)$ & 0.545 & 0.545 & 480k & $(19,48)$ & 0.267 & 0.267 \\
\bottomrule
\end{tabular}
\end{table}

\begin{table}[h!]
\centering
\small
\caption{Difference to the best tested configuration at each budget, averaged over five budgets. The horizon rule is calibrated on one domain and evaluated on the other. Lower is better.}
\vspace{1mm}
\label{tab:harness-budget-regret}
\begin{tabular}{@{}lccc@{}}
\toprule
Calibration $\to$ evaluation & Horizon rule & Longest horizon & Largest batch \\
\midrule
TBLite $\to$ DeepSWE & \textbf{0.003} & 0.029 & 0.031 \\
DeepSWE $\to$ TBLite & \textbf{0.038} & 0.054 & 0.082 \\
\bottomrule
\end{tabular}
\end{table}

\paragraph{Transfer from TBLite to DeepSWE} We reuse the starting artifact that performed best in the TBLite study. We also use the TBLite experiments to set a target horizon of eight steps, about two-thirds of $\rho$. We transfer this fraction to DeepSWE using its pilot token counts, giving $L^\star \approx (2/3)\rho \approx 51$ steps. At each context budget, we select the tested batch--horizon pair closest to this target, without using DeepSWE performance to make that choice. Table~\ref{tab:deep-swe-horizon-transfer} compares the resulting harness with the best score among the three configurations tested at that budget. We also calibrate the horizon rule on DeepSWE and evaluate it on TBLite (Appendix~\ref{sec:app:harness:budget}).

\paragraph{Results} Calibrating on TBLite and transferring the choice of $L$ to DeepSWE produces a harness within 0.008 of the best tested score at every DeepSWE budget level (Table~\ref{tab:deep-swe-horizon-transfer}). Calibrating on DeepSWE and evaluating on TBLite gives less consistent results: at 65k tokens, a 31-step horizon scores 0.626, compared with 0.441 for the selected eight-step horizon. However, whether calibrated on TBLite or on DeepSWE, the transferred choice has a smaller difference to the best tested score than choosing either the longest available horizon or the largest available batch (Table~\ref{tab:harness-budget-regret}). When calibrated on TBLite and evaluated on DeepSWE, its difference to the best configuration is 8.8 times smaller than that of the longest horizon, computed before rounding. The longest horizon gives the best score at only 2 of the 10 experiment setups (five budgets in each domain). The TBLite-to-DeepSWE result shows that a choice in the design of learning context can transfer after adjusting for the target domain's token costs.

\vspace{-1mm}
\section{Conclusion and Discussion} \label{sec:conclusion}
\vspace{-1mm}

We analyzed the difficulty of setting up a \textit{learning loop} for generative optimization, focusing on the \textit{starting artifact}, \textit{credit horizon}, and \textit{experience batching}. Individual ablations highlighted sensitivity to each decision and motivated a unified harness optimization experiment applying all three factors. This experiment also examined how to choose a credit horizon under a shared token budget, using a rule calibrated on TBLite and tested on DeepSWE.

Generative optimization has other degrees of freedom that we did not ablate, such as the design of the feedback oracle, the choice of LLM as the optimizer, and specific optimization procedures.
We instead sought to isolate learning-loop design choices that are often treated as implementation details but can materially change outcomes.
Interestingly, these choices closely parallel long-studied concepts in machine learning.
Viewed through this lens, we conjecture that with sustained research interests, generative optimization may eventually admit robust ``defaults'' that enable broad adoption: just as Transformers~\citep{vaswani2017attention} provided a broadly useful inductive bias for sequence modeling, we may discover \textit{starting artifacts} for agents that are broadly optimizable across tasks; and just as Adam~\citep{kingma2014adam} works well across a wide range of neural architectures, we may discover robust ways to structure the \textit{learning context} --- what traces to include, truncate, and batch --- that transfer across agent designs and domains.

\bibliography{references,neurips_sections_v2/rebuttal_references}
\bibliographystyle{icml2026}

\newpage
\appendix

\newcommand{\appsectiontoc}[2]{%
    \noindent\hyperref[#1]{\textbf{\ref*{#1}\quad #2}}\dotfill\pageref{#1}\par
}
\newcommand{\appsubsectiontoc}[2]{%
    \noindent\hspace*{1em}\hyperref[#1]{\ref*{#1}\quad #2}\dotfill\pageref{#1}\par
}

\appsectiontoc{sec:app:mlagent}{MLAgentBench Details}
\appsubsectiontoc{sec:app:mlagent:setup}{Experimental setup}
\appsubsectiontoc{sec:app:mlagent:workflow}{Agent design details}
\appsubsectiontoc{sec:mlagent_feedback}{Feedback design details}
\appsubsectiontoc{sec:app:mlagent:learning_dynamics}{Learning dynamics and meta-overfitting}
\appsubsectiontoc{sec:app:mlagent-code}{Representative learned ML pipeline examples}

\medskip
\appsectiontoc{sec:app:atari}{Atari Details}
\appsubsectiontoc{sec:app:atari:setup}{Experimental setup}
\appsubsectiontoc{sec:app:atari-agent-design}{Agent design details}
\appsubsectiontoc{sec:atari_feedback}{Feedback design details}
\appsubsectiontoc{sec:app:atari:horizon}{Credit horizon and rollout design}
\appsubsectiontoc{sec:app:deep-rl-result}{Object-centric deep RL baselines}
\appsubsectiontoc{sec:app:atari-code}{Representative learned artifact examples}

\medskip
\appsectiontoc{sec:app:mini-batch-learning}{BigBench Extra Hard Details}
\appsubsectiontoc{sec:app:bbeh_setup}{Experimental setup}
\appsubsectiontoc{sec:app:bbeh_agent}{Agent design details}
\appsubsectiontoc{sec:app:bbeh_feedback}{Feedback design details}
\appsubsectiontoc{sec:app:batchify}{Batchify operator details}
\appsubsectiontoc{sec:app:bbeh_observations}{Task-specific observations}
\appsubsectiontoc{sec:app:bbeh_dspy_langgraph}{Additional Framework Training Details}

\medskip
\appsectiontoc{sec:app:harness}{Harness Optimization Details}
\appsubsectiontoc{sec:app:harness:sweep}{Three-factor sweep}
\appsubsectiontoc{sec:app:harness:budget}{Fixed-budget sweeps}

\medskip
\appsectiontoc{sec:algorithm}{Building a Learning Loop: Extended Discussion}
\appsubsectiontoc{sec:app:formalizing_learning_loop}{The OPTO framework}
\appsubsectiontoc{sec:app:workflow_graphs}{Workflow graphs and learning graphs}
\appsubsectiontoc{sec:app:graph_templates}{Building learning graphs by template}
\appsubsectiontoc{sec:app:starting_artifact}{The choice of starting artifact}
\appsubsectiontoc{sec:app:credit_horizon}{The choice of credit horizon}
\appsubsectiontoc{sec:app:experience_batching}{The choice of experience batching}
\appsubsectiontoc{sec:app:feedback_design}{Additional considerations: feedback design}

\medskip
\appsectiontoc{sec:app:discussion}{Further Discussion and Disclosures}
\appsubsectiontoc{sec:app:agents_scope}{Agents, agentic systems, and the scope of this paper}
\appsubsectiontoc{sec:app:disclosures}{LLM use disclosures and access cards}
\appsubsectiontoc{sec:app:licenses}{Existing Assets, Licenses, and Terms of Use}

\medskip
\appsectiontoc{sec:app:code_listings}{Optimized code examples}
\appsubsectiontoc{sec:app:code_listings:mlagent}{MLAgentBench code examples}
\appsubsectiontoc{sec:app:code_listings:atari}{Atari code examples}

\medskip

\setcounter{figure}{0}
\renewcommand{\thefigure}{A\arabic{figure}}
\setcounter{table}{0}
\renewcommand{\thetable}{A\arabic{table}}

\section{MLAgentBench Details}
\label{sec:app:mlagent}

\subsection{Experimental Setup}
\label{sec:app:mlagent:setup}

We compare against ResearchAgent~\citep{huang2023mlagentbench}. For this baseline, we also pre-download the datasets and ensure that the agent can produce valid Kaggle submission files. Both systems use the same LLM API, and we report average and best submission scores over five trials.

The MLAgentBench experiments instantiate the starting-artifact problem on a task that asks the LLM optimizer to improve an end-to-end machine learning pipeline. We consider two Kaggle tasks, Housing Price and Spaceship Titanic, and compare two initialization schemes: a one-function workflow and a more modular many-function workflow. In both cases, the optimizer receives the same task description and the same high-level implementation hints through docstrings; what differs is the decomposition of the initial workflow.

Our setup slightly differs from the original MLAgentBench setup~\citep{huang2023mlagentbench}, where the LLM agent is additionally asked to download the dataset and generate the submission file. We instead download the dataset outside the agent and generate the submission file after optimization completes. The difference is not consequential for our purposes because we want to isolate the effect of different starting artifacts on the optimization outcome.

\begin{table}[htbp]
\centering
\small
\begin{tabular}{@{}llcc@{}}
\toprule
       &                & Housing Price & Spaceship Titanic \\
       &                & RMSE ($\downarrow$) & Accuracy (\%, $\uparrow$) \\ \midrule
\multicolumn{4}{c}{ResearchAgent~\citep{huang2023mlagentbench}} \\ \midrule
Average        & --             & 0.149         & 78.17 \\
Best           & --             & 0.145         & 79.84 \\ \midrule
\multicolumn{4}{c}{Learned ML Pipeline (Ours)} \\ \midrule
Average        & One         & \textbf{0.135} & 79.65 \\
               & Multi           & 0.147         & \textbf{79.69} \\ \cmidrule(l){2-4}
Best           & One         & \textbf{0.129} & 80.00 \\
               & Multi           & 0.141         & \textbf{80.43} \\
\bottomrule
\end{tabular}
\caption{Kaggle submission scores over five trials. One and Multi denote the one-function and many-function pipelines. Lower Housing Price RMSE and higher Spaceship Titanic accuracy are better. All conditions use the same LLM API.}
\label{tab:mlagent_main_result}
\end{table}

\paragraph{Submission scores} Table~\ref{tab:mlagent_main_result} reports average and best submission scores for both initializations and ResearchAgent. On Housing Price, the one-function pipeline has lower RMSE than the many-function pipeline in both average and best performance. On Spaceship Titanic, the mean accuracies are close (79.65\% versus 79.69\%), while the best many-function pipeline reaches 80.43\% compared with 80.00\% for the one-function pipeline.

\subsection{Agent Design Details}
\label{sec:app:mlagent:workflow}

The learned ML pipeline shares a similar design for both tasks with modular components for different steps of the machine learning pipeline. In the many-function initialization, the workflow is broken into components such as preprocessing, feature selection, ensemble construction, training, and prediction. This decomposition makes explicit which parts of the pipeline the optimizer is allowed to revise and gives the optimizer a more structured hypothesis class than a single monolithic \texttt{train\_model} function. The schematic comparison below is adapted from the Housing Price implementations
\footnote{\anonurl{https://anonymous.4open.science/r/data-science-agent-6D5E/house-prices/multi_function.py}{https://github.com/AbhinavAkkiraju/data-science-agent/blob/main/house-prices/multi_function.py}}
though for both tasks, we use the same implementation. 

\begin{figure}[ht]
\centering
\begin{subfigure}[t]{0.48\linewidth}
\small
\begin{lstlisting}[language=Python]
@trace.model
class Pipeline:
    def __call__(self, x, y=None, test_data=None):
        # [Workflow structure / specs]
        # One editable function through the pipeline
        return self.train_model(x, y, test_data)

    @trace.bundle(trainable=True)
    def train_model(self, x, y=None, test_data=None):
        """
        [Program documentation]
        Task guidance, constraints,
        and modeling hints live here.

        In the MLAgentBench example, 
        the docstring for this function 
        is a concatenation of all docstrings 
        for the many-function initialization.
        """
        # [Initial implementation]
        # Starter code for the full pipeline
        ...
\end{lstlisting}
\caption{One-function initialization.}
\end{subfigure}\hfill
\begin{subfigure}[t]{0.48\linewidth}
\small
\begin{lstlisting}[language=Python]
@trace.model
class Pipeline:
    def __call__(self, x, y=None, test_data=None):
        # [Workflow structure / specs]
        # Explicit interfaces between multiple editable functions
        x = self.preprocess(x)
        z = self.select_features(x)
        m = self.train_model(z, y)
        return self.predict(m, z)

    @trace.bundle(trainable=True)
    def preprocess(self, x):
        """[Program documentation]"""
        ...  # [Initial implementation]

    @trace.bundle(trainable=True)
    def select_features(self, x):
        """[Program documentation]"""
        ...  # [Initial implementation]

    @trace.bundle(trainable=True)
    def train_model(self, z, y):
        """[Program documentation]"""
        ...  # [Initial implementation]

    @trace.bundle(trainable=True)
    def predict(self, m, z):
        """[Program documentation]"""
        ...  # [Initial implementation]
\end{lstlisting}
\caption{Many-function initialization.}
\end{subfigure}
\caption{\textbf{Trace program for the two MLAgentBench initializations.} The one-function design exposes the full pipeline through a single trainable function, whereas the many-function design exposes preprocessing, feature selection, model training, and prediction as separate trainable components.}
\label{fig:mlagent-init-skeleton}
\end{figure}

\subsection{Feedback Design Details}
\label{sec:mlagent_feedback}

We additionally experiment with improvement-style feedback. When the model fails to improve the task-specific validation metric relative to the previous step, we append an improvement suggestion to the feedback string.

We provide task-specific feedback instructions when the agent reaches different performance levels. Figure~\ref{fig:mlagent-feedback-template} shows the feedback templates used for Spaceship Titanic and Housing Price. Table~\ref{tab:mlagent-feedback-structure} summarizes the staged suggestive feedback used for both tasks. This feedback design is intended to give the optimizer more direction than a bare validation score while still leaving room for nontrivial revisions.

\begin{figure}[ht]
    \centering
\begin{subfigure}[t]{0.48\linewidth}
\centering
\begin{AIbox}{Spaceship Titanic Feedback Template}
\small
\textbf{Epoch} [epoch]\textbf{/20}\par
\medskip
\textbf{Accuracy:} [val\_accuracy]\par
\textbf{F1:} [val\_f1]\par
\textbf{Precision:} [val\_precision]\par
\textbf{Recall:} [val\_recall]\par
\medskip
[SUGGESTION]
\end{AIbox}
\caption{Spaceship Titanic.}
\end{subfigure}\hfill
\begin{subfigure}[t]{0.48\linewidth}
\centering
\begin{AIbox}{Housing Price Feedback Template}
\small
\textbf{Epoch} [epoch]\textbf{/20}\par
\medskip
\textbf{RMSE:} [val\_rmse]\par
\textbf{MAE:} [val\_mae]\par
\textbf{r\textsuperscript{2}:} [val\_r2]\par
\medskip
[SUGGESTION]
\end{AIbox}
\caption{Housing Price.}
\end{subfigure}
\caption{Feedback templates used for the learned ML pipeline on the Spaceship Titanic and Housing Price tasks.}
\label{fig:mlagent-feedback-template}
\end{figure}

\newcommand{\mlfeedbackhl}[1]{%
  \begingroup
  \setlength{\fboxsep}{1pt}%
  \colorbox{paramyellow}{\strut #1}%
  \endgroup
}

\begin{table*}[ht]
\centering

\begin{subtable}[t]{0.48\textwidth}
\centering
\renewcommand{\arraystretch}{1.25}
\small
\resizebox{\linewidth}{!}{%
\begin{tabular}{p{2.7cm}p{4.8cm}}
\toprule
\textbf{Validation F1} & \textbf{Suggestive Feedback} \\
\midrule
Val F1 $<$ 0.5 
& \raggedright ``\mlfeedbackhl{Model performance is poor}. \mlfeedbackhl{Try} better feature engineering and preprocessing.'' \tabularnewline
\midrule
0.5 $\leq$ Val F1 $<$ 0.7 
& \raggedright ``\mlfeedbackhl{Model is showing promise but needs} \newline 
\mlfeedbackhl{improvement}. Consider \mlfeedbackhl{class balancing techniques}.'' \tabularnewline
\midrule
0.7 $\leq$ Val F1 $<$ 0.8 
& \raggedright ``\mlfeedbackhl{Model is performing well}. \newline 
\mlfeedbackhl{Fine-tune hyperparameters for} \newline 
\mlfeedbackhl{further improvements}.'' \tabularnewline
\midrule
Val F1 $\geq$ 0.8 
& \raggedright ``\mlfeedbackhl{Excellent performance!} Focus on \mlfeedbackhl{preventing overfitting}.'' \tabularnewline
\bottomrule
\end{tabular}%
}
\caption{Spaceship Titanic feedback.}
\label{tab:mlagent-spaceship-titanic-feedback-structure}
\end{subtable}
\hfill
\begin{subtable}[t]{0.48\textwidth}
\centering
\renewcommand{\arraystretch}{1.25}
\small
\resizebox{\linewidth}{!}{%
\begin{tabular}{p{2.7cm}p{4.8cm}}
\toprule
\textbf{Validation $r^2$} & \textbf{Suggestive Feedback} \\
\midrule
$r^2 \leq 0$ 
& \raggedright ``\mlfeedbackhl{Model is performing worse than} \newline 
\mlfeedbackhl{baseline}. \mlfeedbackhl{Focus on} better feature engineering and selection.'' \tabularnewline
\midrule
$0 < r^2 < 0.5$ 
& \raggedright ``\mlfeedbackhl{Model has poor predictive power}. Try \mlfeedbackhl{more advanced preprocessing} \newline 
\mlfeedbackhl{or different algorithms}.'' \tabularnewline
\midrule
$0.5 \leq r^2 < 0.7$ 
& \raggedright ``\mlfeedbackhl{Model is improving but still has} \newline 
\mlfeedbackhl{room for growth}. Consider \mlfeedbackhl{feature interactions}.'' \tabularnewline
\midrule
$r^2 \geq 0.7$ 
& \raggedright ``\mlfeedbackhl{Model is performing well}. \newline 
\mlfeedbackhl{Fine-tune hyperparameters} for further improvements.'' \tabularnewline
\bottomrule 
\end{tabular}%
}
\caption{Housing Price feedback.}
\label{tab:mlagent-housing-feedback-structure}
\end{subtable}

\caption{Staged feedback templates for Spaceship Titanic and Housing Price. Highlighted phrases mark the parts that differ across feedback levels within each task.}
\label{tab:mlagent-feedback-structure}
\end{table*}

\subsection{Learning Dynamics and Meta-Overfitting}
\label{sec:app:mlagent:learning_dynamics}

\paragraph{Train/Validation Data Splitting} We create a train-validation split outside the agent and use the validation metric as the optimization signal. Kaggle test submissions are reserved for final external evaluation and are never used during optimization. We use OptoPrime~\citep{cheng2024trace} as the generative optimizer and evaluate each configuration across multiple trials.
Kaggle does not permit enough test submissions for the test set to be used as an optimization signal, so the held-out split within the available training data becomes the reward source for generative optimization. We provide 80\% of the original training data to the agent and reserve the remaining 20\% as the internal validation split used for feedback and checkpoint selection.

\paragraph{Optimization Details} Concretely, the outer optimization loop is identical between the one-function and many-function initializations. In both cases, we fix a train-validation split outside the agent, run OptoPrime for 20 optimization steps with memory size 5, compute validation metrics after each step, and convert those metrics into a natural-language feedback string before calling the optimizer update. For the Spaceship Titanic task shown here, staged feedback thresholds and checkpoint selection both use validation F1; accuracy, precision, and recall are included in the feedback string as auxiliary diagnostics. The checkpoint with the best task-specific internal validation metric is then used to generate the final Kaggle submission. Thus, the main difference between the two initializations is not the outer training protocol, but the editable part of the program exposed to the optimizer.

\begin{figure}
    \centering
    \includegraphics[width=0.8\linewidth]{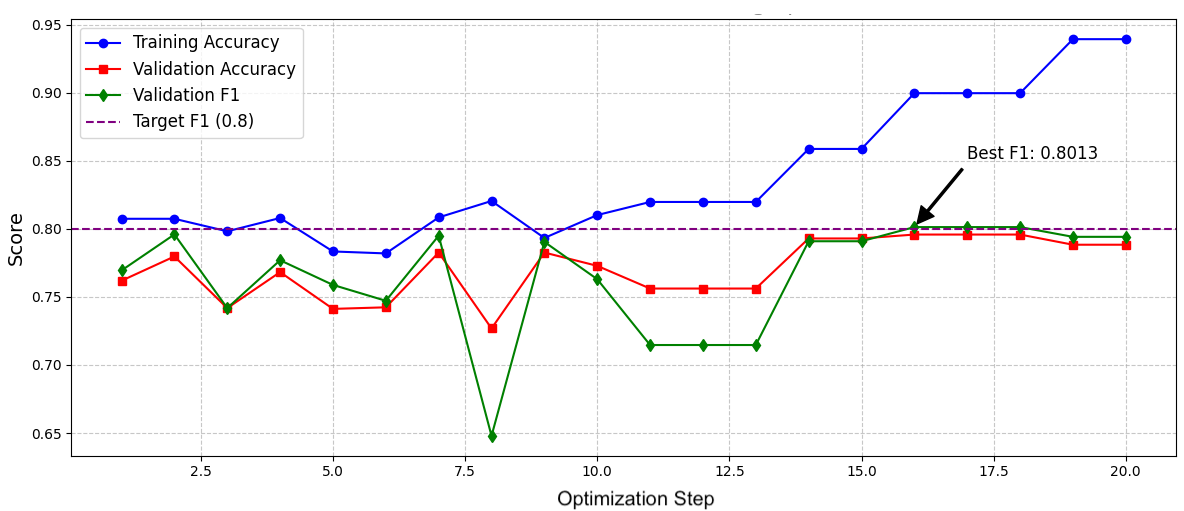}
    \caption{\textbf{One Optimization Trajectory of LLM Optimizer on the ML Pipeline.} We show a single optimization trajectory from the Spaceship Titanic dataset. The feedback string reports Accuracy, F1 score, Precision, and Recall, but both staged feedback and checkpoint selection are based on validation F1. Each dot (step) represents a full execution of the training pipeline code written by the LLM optimizer with a fully trained model. We see that LLM optimizer tends to overfit to the training data and exhibits classical overfitting behavior even without numerical gradient descent.}
    \label{fig:mlagent-titanic-learning-graph}
\end{figure}

Figure~\ref{fig:mlagent-titanic-learning-graph} shows the learning progress of \textit{one trial} of the Spaceship Titanic task. Although it resembles ordinary model overfitting at first glance, the figure reflects a different phenomenon. At each optimization step, the agent produces a \emph{new fully trained model} from scratch. What increases over time is the optimizer's tendency to discover pipeline code revisions that fit the training split at the expense of generalization. We refer to this as \emph{meta-overfitting}: the generative optimizer learns to make workflow revisions that improve the immediate validation-driven objective while drifting toward brittle pipelines.

\paragraph{Starting-artifact comparison} On both tasks,
our learned ML pipeline improves the average leaderboard percentile over ResearchAgent~\citep{huang2023mlagentbench} by 2.1--22.4 percentage points,
depending on the task and initialization,
and the best machine learning model produced by the learned ML pipeline surpasses 86.6\% of leaderboard submissions. 
We notice a difference between our two initialization options.
In Figure~\ref{fig:spaceship_one_v_multi},
for the Spaceship Titanic dataset, asking the LLM optimizer to implement and modify a single function (\texttt{train\_model})
yields a weaker best pipeline than implementing and modifying a set of functions, while the mean accuracies are nearly identical (79.65\% versus 79.69\%).
Over 5 trials, if we look at the best pipeline generated under these two initial conditions,
we see a large contrast, with one initial system configuration (one-function) surpassing 72.7\% of leaderboard submissions,
while the other configuration (many-function) surpasses 86.6\%.
However, for the Housing Price dataset, the observed ordering is flipped.
The one-function initial system configuration resulted in the best ML pipeline that produced a model that surpassed 75.6\% of leaderboard submissions, while the many-function initial system configuration only surpassed 54.6\% of submissions.
Here, the difference is noticeable both in average leaderboard percentile (64.1\% versus 43.8\%) and best pipeline across runs.
We show some examples of the learned ML pipeline code in Figure~\ref{fig:mlagent-spaceship-final-agent-1}.

\subsection{Representative Learned ML Pipeline Examples}
\label{sec:app:mlagent-code}

We provide a snippet of the learned ML pipeline code to show how the LLM optimizer writes specific preprocessing logic (Figure~\ref{fig:mlagent-spaceship-final-agent-1}). The full implementation is publicly available at 
\anonurl{https://anonymous.4open.science/r/data-science-agent-6D5E}{https://github.com/AbhinavAkkiraju/data-science-agent}.

\setcounter{figure}{0}
\renewcommand{\thefigure}{B\arabic{figure}}
\setcounter{table}{0}
\renewcommand{\thetable}{B\arabic{table}}

\section{Atari Details}
\label{sec:app:atari}

\subsection{LLM Optimizer Experimental Setup}
\label{sec:app:atari:setup}

We run five trials for each game under both credit-horizon conditions. Per-game optimizer steps and evaluation lengths are reported in Table~\ref{tab:atari-training-configs}.

The per-game training configuration is reported in Table~\ref{tab:atari-training-configs}. Across all eight games, we use frame skipping to shorten the effective horizon and set the sticky-action probability to $0.0$, since in our experiments removing sticky actions led to more stable optimization of the learned policy code.

We generate data on-the-fly using object-centric Atari Environments (OCAtari)~\citep{delfosse2024ocatari}\footnote{\url{https://github.com/k4ntz/OC_Atari}}, a wrapper around the Gymnasium API~\citep{towers2024gymnasium}. Instead of raw pixels, the learned agent receives a structured dictionary of objects at each timestep. This representation makes the revised code more interpretable and gives the optimizer direct access to semantically meaningful state variables. The full eight-game set is Pong, Breakout, Space Invaders, Freeway, Asterix, Enduro, Q*bert, and Seaquest. To keep the appendix compact, the screenshots, feedback templates, and code examples below focus on Pong, Breakout, and Space Invaders as representative games.

The OCAtari wrapper returns game-specific dictionaries. As representative examples, Pong exposes a compact state with \texttt{Player}, \texttt{Ball}, and \texttt{Enemy}. Breakout additionally groups remaining bricks into row-wise lists \texttt{RB/OB/YB/GB/AB/BB} and records \texttt{lives}. 

\paragraph{Pong} In Pong, the player controls a paddle on the right side of the screen to deflect the ball into the enemy's goal. The player scores a point if the enemy misses the ball. The game ends when one side scores 21 points.

\paragraph{Breakout} In Breakout, the player moves a bottom paddle horizontally to deflect a ball that scores against brick walls upon contact. The brick wall consists of six rows of different colored bricks, with higher bricks worth more points. Hitting higher bricks increases ball speed and therefore the difficulty of recovery. The player wins after scoring 864 points and loses a life when the ball drops out of range.

\paragraph{Space Invaders} In Space Invaders, the player controls a cannon at the bottom of the screen and must move left or right while firing at descending aliens. Enemy projectiles, shield usage, and firing cadence make the action-value of any single move more dependent on longer-term state evolution than in Pong or Breakout.

Reset handling is also game-specific in the code. As representative examples, Breakout automatically applies \texttt{FIRE} after reset so that each rollout starts with the ball in play, whereas Pong and Space Invaders use a standard reset.

\begin{table}[ht]
\centering
\small
\resizebox{\linewidth}{!}{%
\begin{tabular}{ll}
\toprule
\textbf{Name} & \textbf{Repo} \\
\midrule
Deep RL Repo (CleanRL) & \anonurl{https://anonymous.4open.science/r/cleanrl_obj_centric-822B}{https://github.com/ameliakuang/cleanrl_obj_centric} \\ 
\midrule
LLM Code Repo & \anonurl{https://anonymous.4open.science/r/LLM-Game-Playing-Agents-09C5}{https://github.com/ameliakuang/LLM-Game-Playing-Agents} \\ 
\midrule 
\midrule
Wandb Log for DQN Training & \ifanonurl Anonymized\else \url{https://wandb.ai/kuangzy-amelia-stanford-university/obj-dqn-5trials-new}\fi \\ 
\midrule
Wandb Log for PPO Training & \ifanonurl Anonymized\else \url{https://wandb.ai/kuangzy-amelia-stanford-university/obj-ppo-5trials-flatten}\fi \\
\bottomrule
\end{tabular}%
}
\vspace{2mm}
\caption{Repositories and experiment logs used for the Atari experiments.}
\label{tab:atari-repos}
\end{table}

\begin{table*}[t]
\centering
\small
\begin{tabular}{lcccc}
\toprule
\textbf{Game} & \textbf{One-Step} & \textbf{Multi-Step} & \textbf{Optimizer Steps} & \textbf{Evaluation Protocol} \\
\midrule
Pong & 1 step & 400 steps & 20 & 10 episodes $\times$ 4000 steps \\
Breakout & 1 step & 300 steps & 30 & 1 episode $\times$ 4000 steps \\
Space Invaders & 1 step & 25 steps & 20 & 1 episode $\times$ 4000 steps \\
Freeway & 1 step & 100 steps & 30 & 3 episodes $\times$ 2500 steps \\
Asterix & 1 step & 100 steps & 30 & 3 episodes $\times$ 20000 steps \\
Enduro & 1 step & 100 steps & 30 & 1 episode $\times$ 5000 steps \\
Q*bert & 1 step & 100 steps & 30 & 1 episode $\times$ 4000 steps \\
Seaquest & 1 step & 100 steps & 30 & 1 episode $\times$ 4000 steps \\
\bottomrule
\end{tabular}
\vspace{2mm}
\caption{Per-game Atari credit-horizon configurations. All games use ALE \texttt{\{env\}-NoFrameskip-v4} environments with action repeat 4, sticky action probability 0.0, optimizer memory size 5, and Claude Sonnet-3.5 via OptoPrime.}
\label{tab:atari-training-configs}
\end{table*}

\begin{figure*}[t]
  \centering

  \begin{subfigure}[t]{0.32\textwidth}
    \centering
    \includegraphics[width=0.49\linewidth]{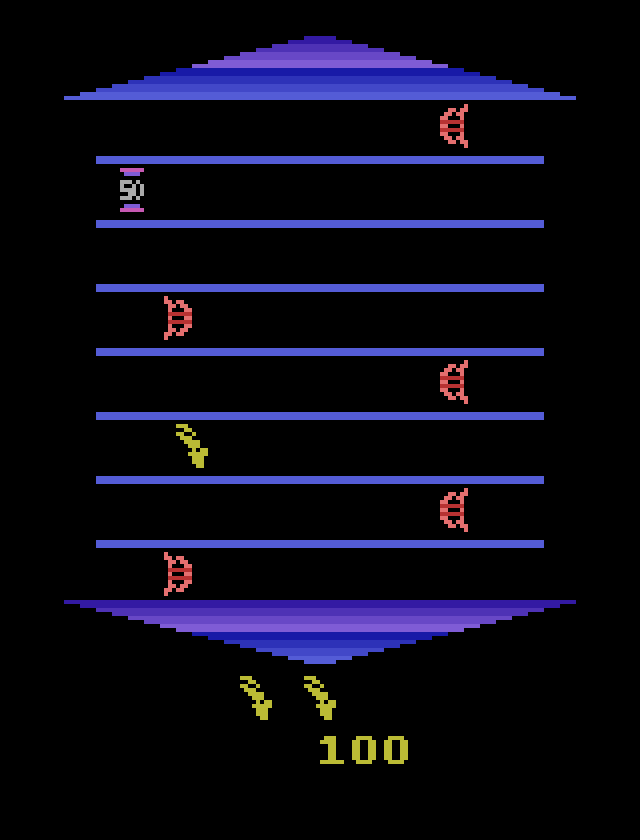}%
    \hfill
    \includegraphics[width=0.49\linewidth]{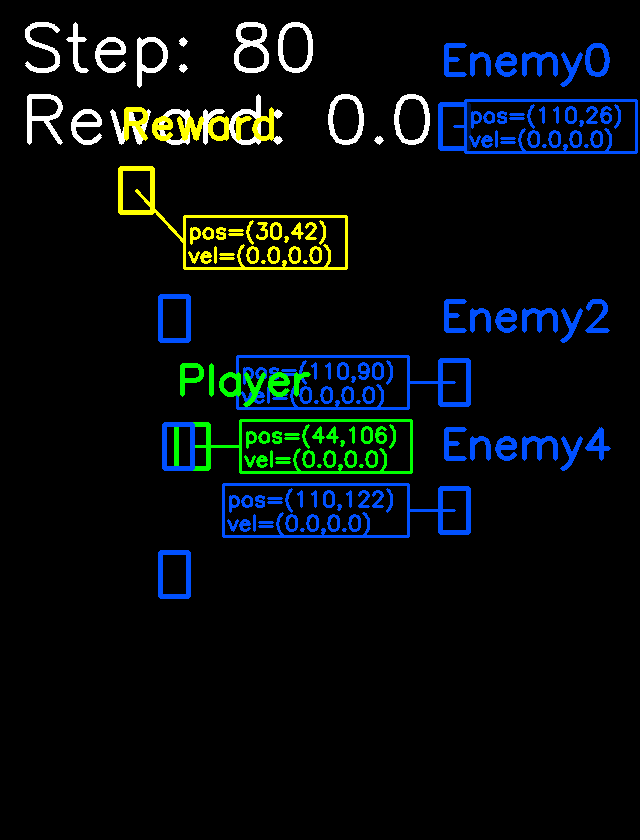}
    \caption{Asterix}
    \label{fig:atari-asterix-screenshot}
  \end{subfigure}
  \hfill
  \begin{subfigure}[t]{0.32\textwidth}
    \centering
    \includegraphics[width=0.49\linewidth]{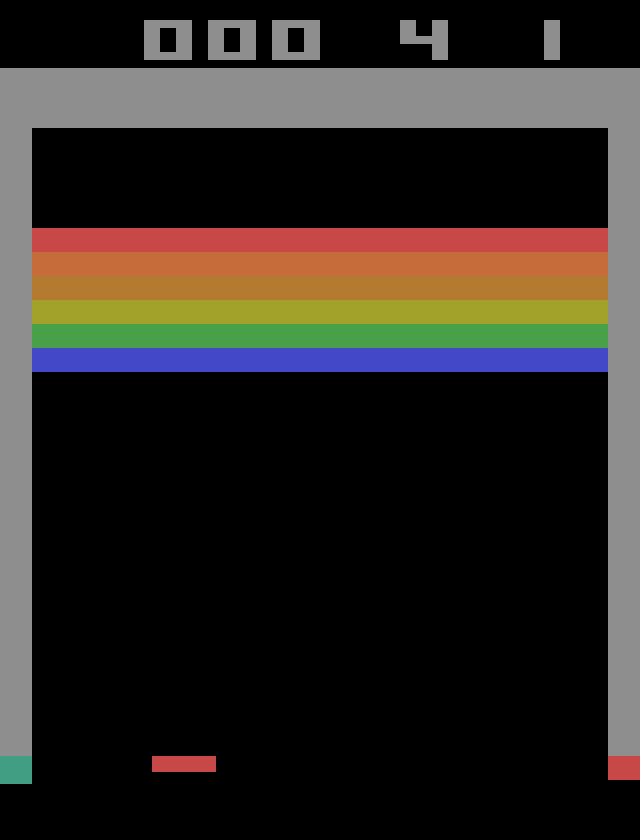}%
    \hfill
    \includegraphics[width=0.49\linewidth]{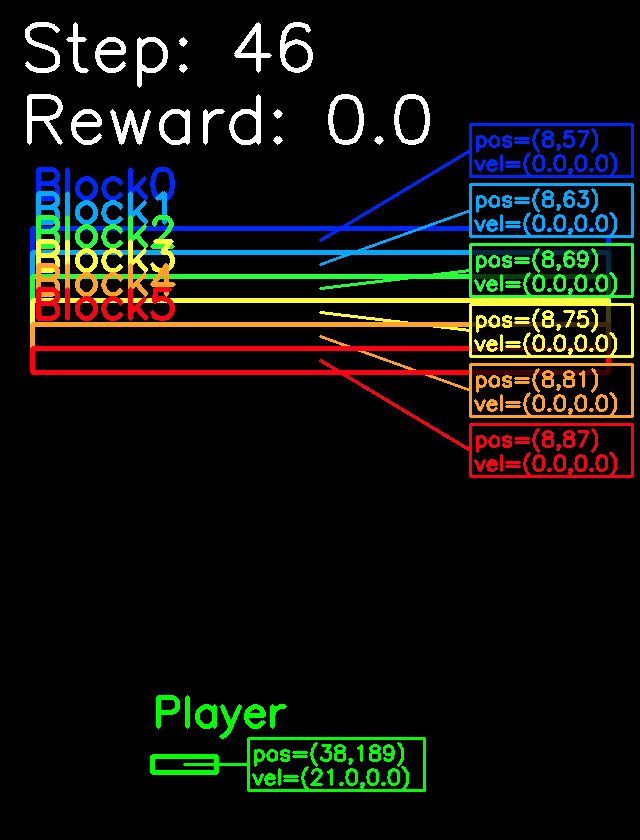}
    \caption{Breakout}
    \label{fig:atari-breakout-screenshot}
  \end{subfigure}
  \hfill
  \begin{subfigure}[t]{0.32\textwidth}
    \centering
    \includegraphics[width=0.49\linewidth]{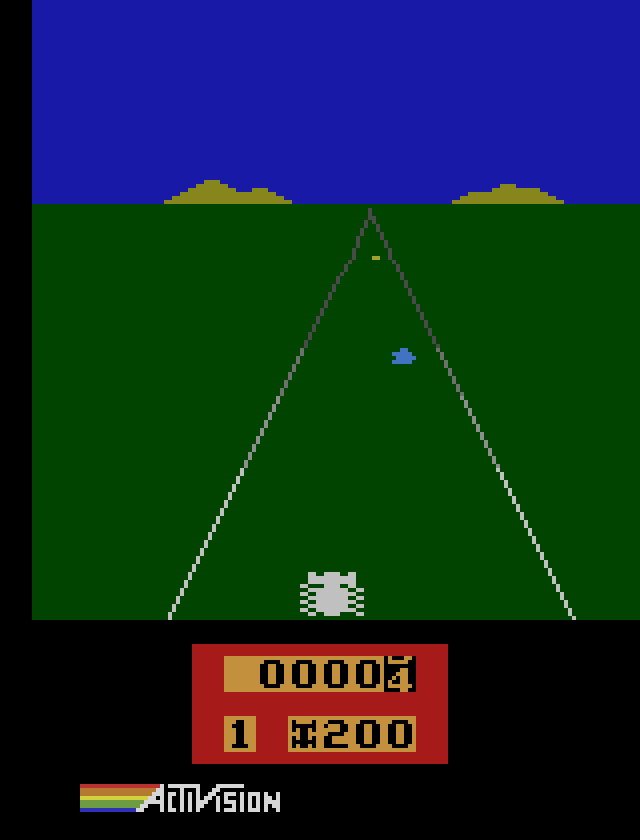}%
    \hfill
    \includegraphics[width=0.49\linewidth]{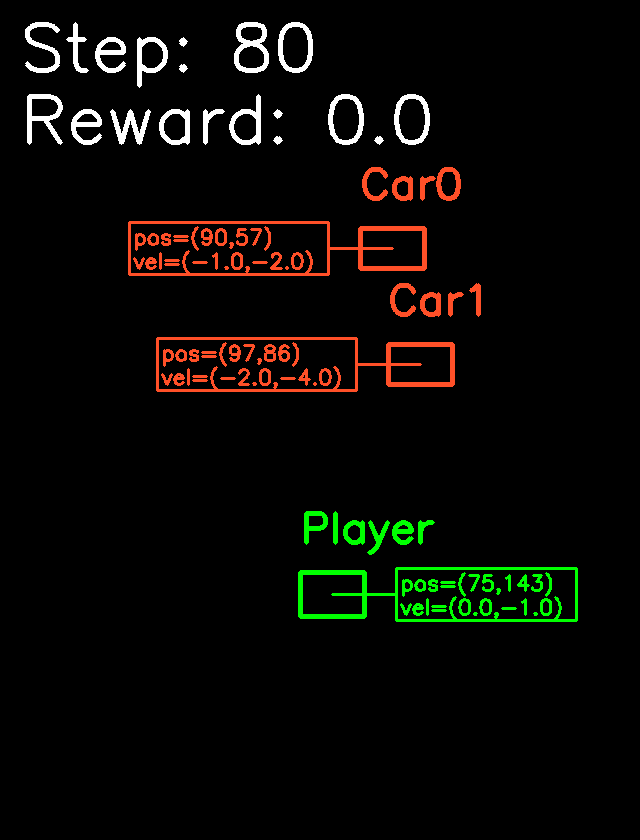}
    \caption{Enduro}
    \label{fig:atari-enduro-screenshot}
  \end{subfigure}

  \vspace{1mm}

  \begin{subfigure}[t]{0.32\textwidth}
    \centering
    \includegraphics[width=0.49\linewidth]{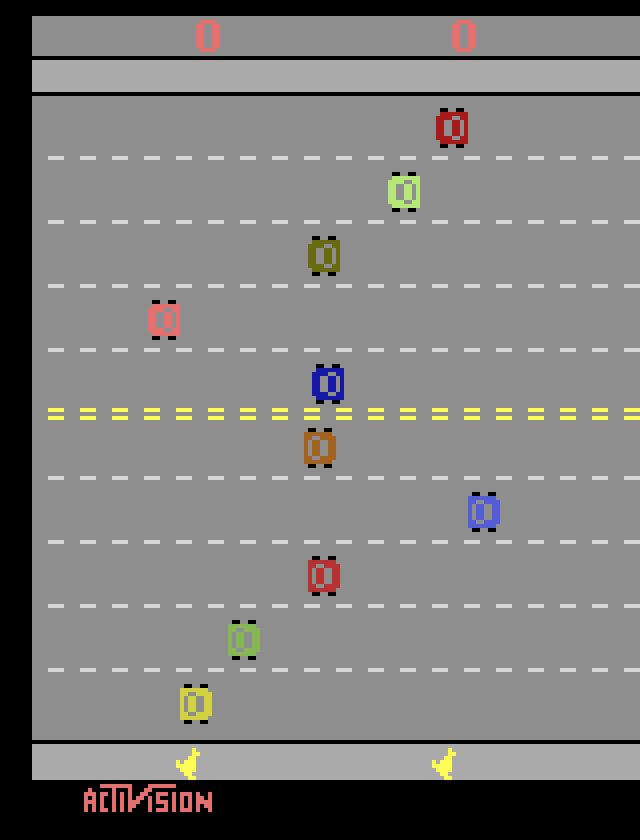}%
    \hfill
    \includegraphics[width=0.49\linewidth]{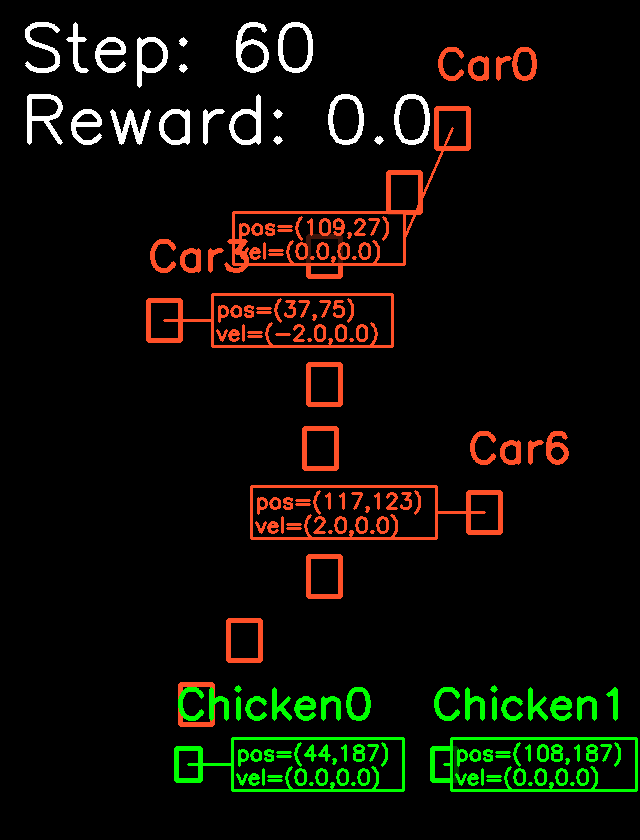}
    \caption{Freeway}
    \label{fig:atari-freeway-screenshot}
  \end{subfigure}
  \hfill
  \begin{subfigure}[t]{0.32\textwidth}
    \centering
    \includegraphics[width=0.49\linewidth]{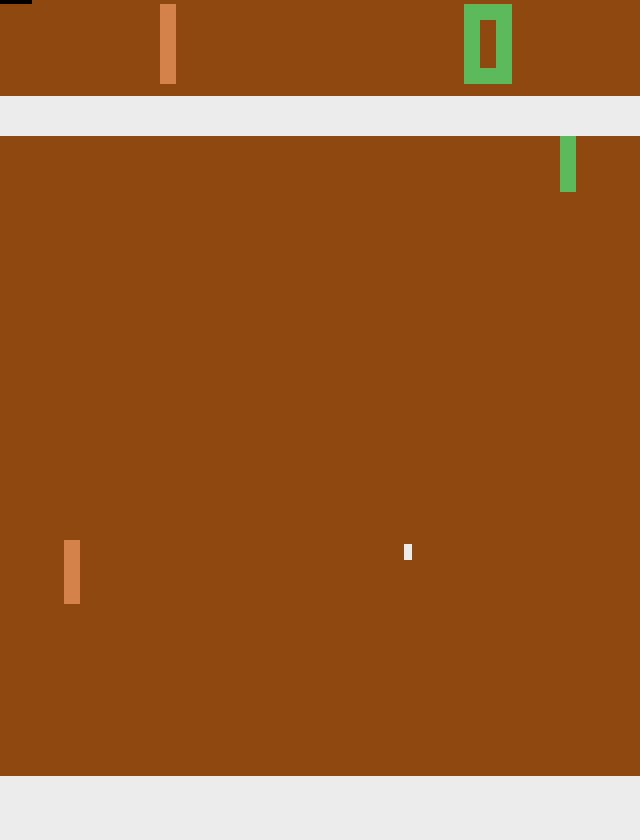}%
    \hfill
    \includegraphics[width=0.49\linewidth]{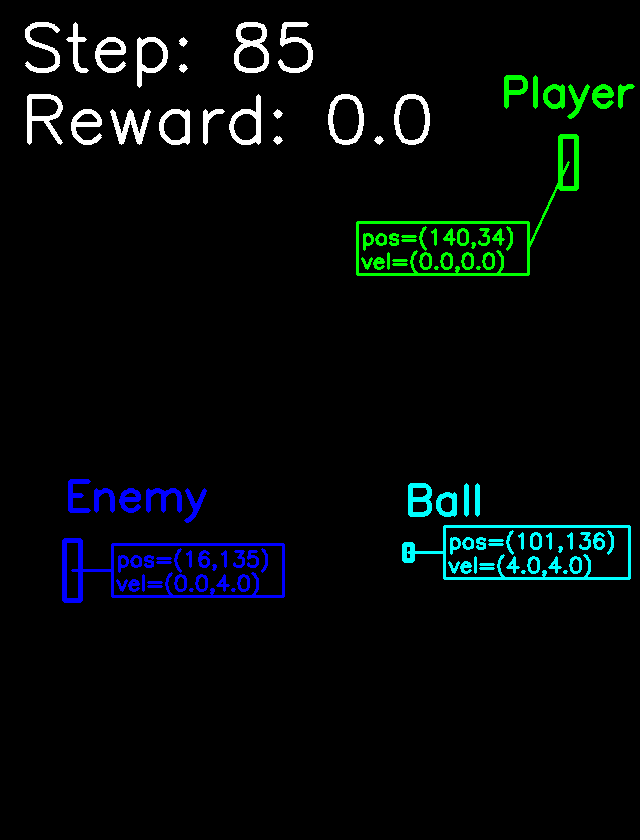}
    \caption{Pong}
    \label{fig:atari-pong-screenshot}
  \end{subfigure}
  \hfill
  \begin{subfigure}[t]{0.32\textwidth}
    \centering
    \includegraphics[width=0.49\linewidth]{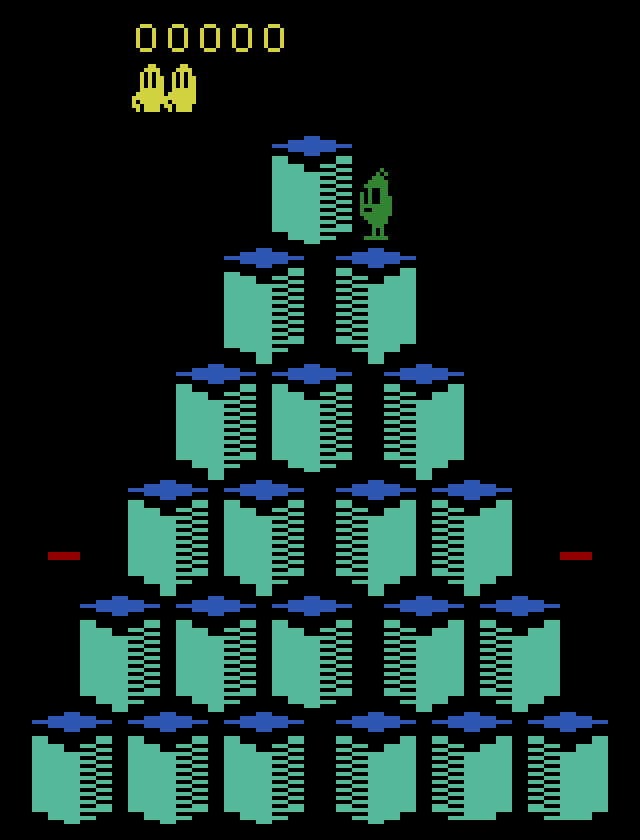}%
    \hfill
    \includegraphics[width=0.49\linewidth]{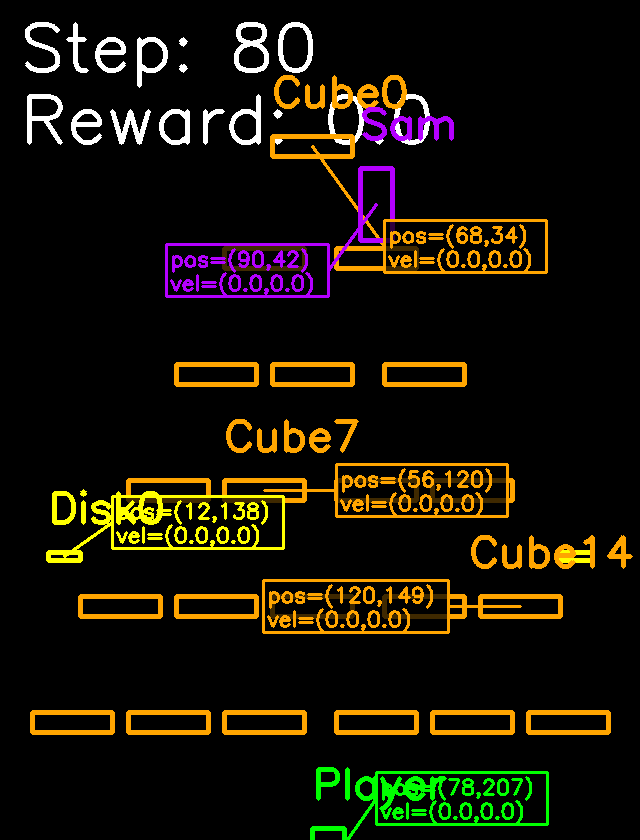}
    \caption{Q*bert}
    \label{fig:atari-qbert-screenshot}
  \end{subfigure}

  \vspace{1mm}

  \makebox[\textwidth][c]{%
    \begin{subfigure}[t]{0.32\textwidth}
      \centering
      \includegraphics[width=0.49\linewidth]{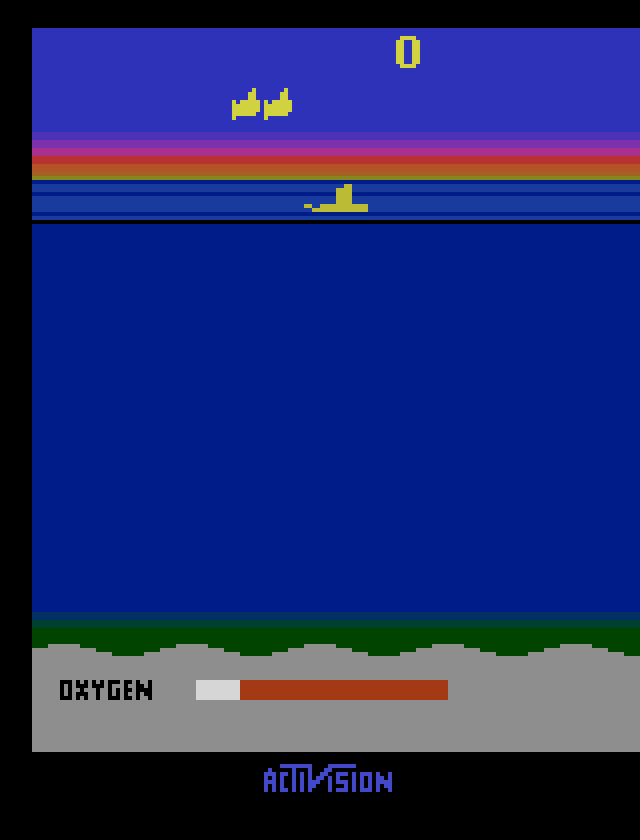}%
      \hfill
      \includegraphics[width=0.49\linewidth]{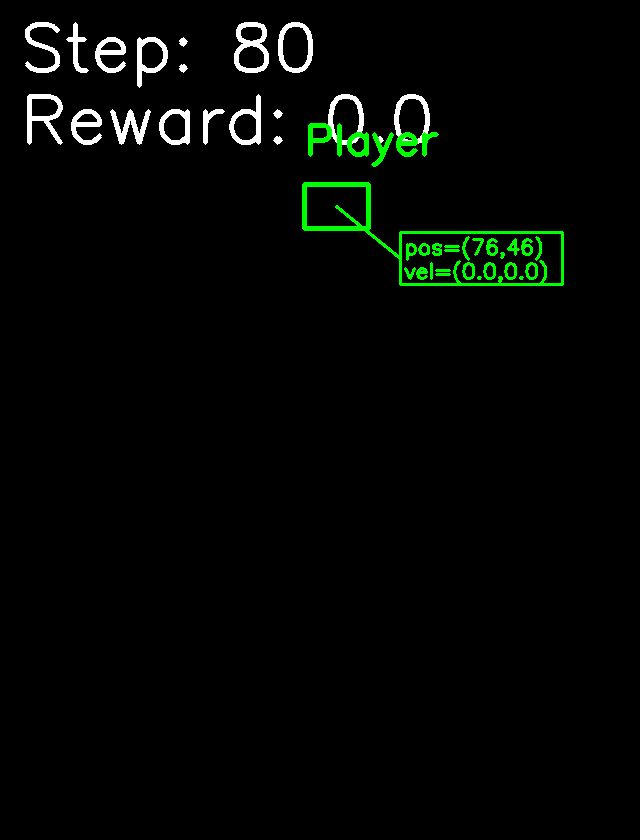}
      \caption{Seaquest}
      \label{fig:atari-seaquest-screenshot}
    \end{subfigure}
    \hspace{0.03\textwidth}
    \begin{subfigure}[t]{0.32\textwidth}
      \centering
      \includegraphics[width=0.49\linewidth]{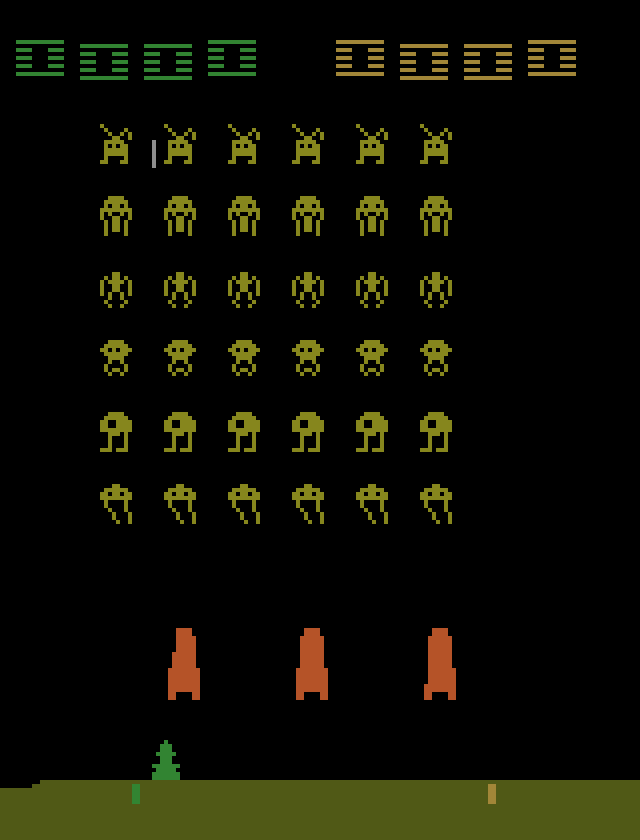}%
      \hfill
      \includegraphics[width=0.49\linewidth]{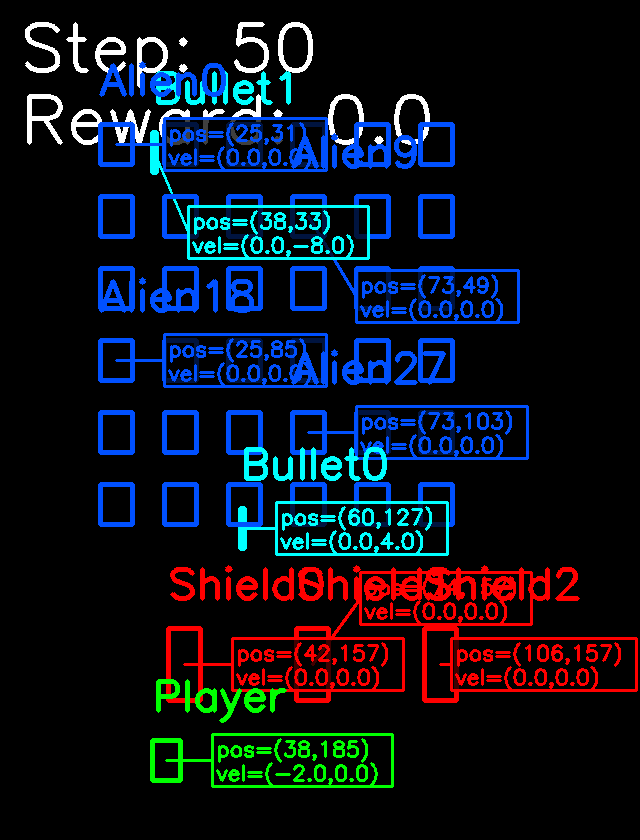}
      \caption{Space Invaders}
      \label{fig:atari-space-invaders-screenshot}
    \end{subfigure}
  }

  \caption{\textbf{OCAtari object annotations across eight Atari games.}
  For each game, the left image shows the raw RGB frame and the right image shows the corresponding OCAtari~\citep{delfosse2024ocatari} annotated frame. These examples illustrate how pixel observations are translated into object-centric state representations across diverse Atari environments.}
  \label{fig:atari-annotated-screenshots}
\end{figure*}

\begin{figure}[ht]
    \centering
\begin{AIbox}{OCAtari Breakout Observation Example}
\small
\texttt{TracedEnv.step.step16 = \{}\\
\texttt{\ \ `Player': \{`x': 99, `y': 189, `w': 16, `h': 4, `dx': 0, `dy': 0\},}\\
\texttt{\ \ `Ball': \{`x': 7, `y': 193, `w': 2, `h': 4, `dx': -4, `dy': 4\},}\\
\texttt{\ \ `RB': [\{`x': 8, `y': 57, `w': 144, `h': 6\}],}\\
\texttt{\ \ `OB': [\{`x': 8, `y': 63, `w': 144, `h': 6\}],}\\
\texttt{\ \ `YB': [\{`x': 8, `y': 69, `w': 144, `h': 6\}],}\\
\texttt{\ \ `GB': [\{`x': 8, `y': 75, `w': 144, `h': 6\}],}\\
\texttt{\ \ `AB': [\{`x': 8, `y': 81, `w': 144, `h': 6\}],}\\
\texttt{\ \ `BB': [\{`x': 8, `y': 87, `w': 144, `h': 6\}],}\\
\texttt{\ \ `lives': 5,}\\
\texttt{\ \ `reward': 0.0}\\
\texttt{\}}
\end{AIbox}
\caption{Example of a returned observation, game information, and reward from Breakout. Note that this is the raw return from OCAtari, not further processed. There is no annotation of what ``RB'', ``OB'', ``YB'', ``GB'', ``AB'', and ``BB'' represent. They are acronyms for the bounding boxes with colors.}
\label{fig:atari-traced-step}
\end{figure}

\subsection{Agent Design Details}
\label{sec:app:atari-agent-design}

The learned Python policy is stateless, taking the structured OCAtari state at time $t$ and returning an action $a_t$. We pass the object dictionary directly without additional manual feature engineering or natural-language annotation.

\begin{figure*}[ht]
  \centering
  \includegraphics[width=\linewidth]{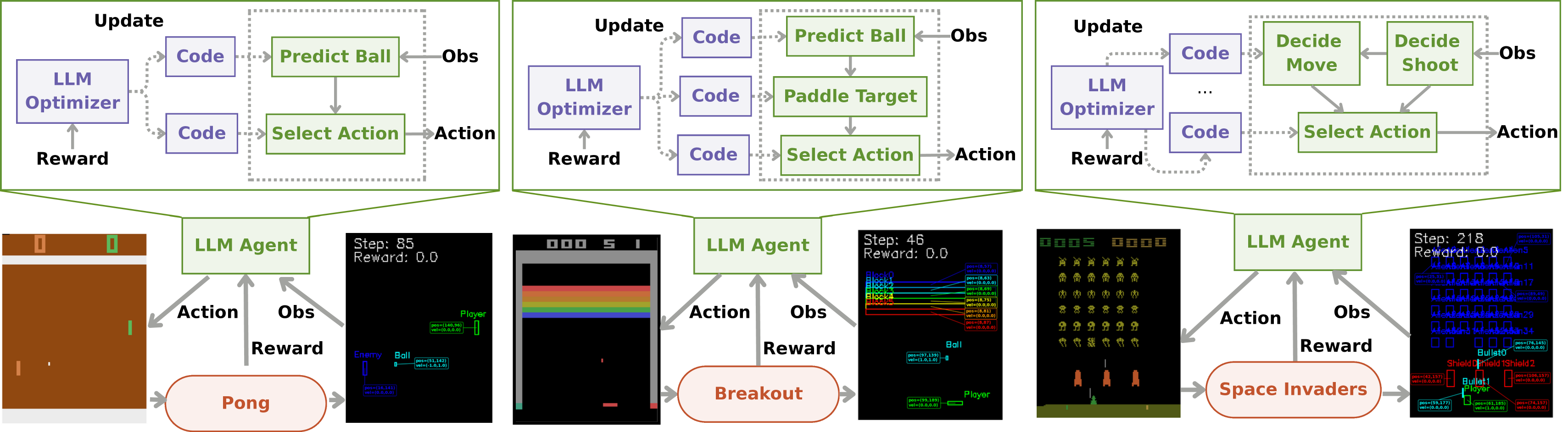}
  \caption{\textbf{Representative initial systems for Atari games}. We show the workflow design of different decision-making program components for three representative Atari game agents. The learned agent receives an object-centric dictionary of game state information and uses Python code to process the observation and output an action.}
  \label{fig:atari-screenshot}
  \vspace{-4mm}
\end{figure*}

\paragraph{Pong} To succeed at Pong, the agent should accurately predict where the ball will intersect the player's paddle plane while accounting for wall bounces. We therefore use a workflow centered on trajectory prediction and paddle control, with trainable functions such as \texttt{predict\_ball\_trajectory()} and \texttt{select\_action()}. The initialized version returns the current ball coordinate and a random movement decision; the learned version uses a more structured geometric heuristic.

\paragraph{Breakout} Breakout also depends on ball trajectory, but the objective is richer because paddle placement affects which bricks the ball can target next. The workflow therefore includes \texttt{predict\_ball\_trajectory()}, \texttt{generate\_paddle\_target()}, and \texttt{select\_paddle\_action()}. The initialized agent is intentionally underspecified so the optimizer must discover both reliable returns and useful targeting behavior.

\paragraph{Space Invaders} For Space Invaders, we separate movement and firing into \texttt{decide\_shoot()}, \texttt{decide\_movement()}, and \texttt{combine\_actions()}. This decomposition makes clear that the game requires concurrent control over attack and defense rather than pure interception of a moving ball.

\subsection{Feedback Design Details}
\label{sec:atari_feedback}

We provide game-specific feedback instructions when the agent reaches different reward regions. The goal is not only to report score, but also to guide the optimizer toward qualitatively better behavior for each game. Here, we explore a type of feedback design similar to traditional reward shaping, where we provide different feedback at different stages of the agent performance. Without the need to constantly have a human-in-the-loop to provide per-step feedback, staged feedback still allows engineers to inject dynamic information and provide granular domain-specific guidance to the LLM. In the actual optimization loop, the optimizer sees both the execution trace, which contains the raw rewards, and an appended template-based natural-language feedback string derived from those rewards. We give representative examples in Tables~\ref{tab:atari-feedback-structure} and~\ref{tab:feedback-structure-space}. Note that the feedback is template-based, and that the performance thresholds (set on the trajectory level) triggering different messages are pre-determined by the engineer.

\newcommand{\feedbackhl}[1]{%
  \begingroup
  \setlength{\fboxsep}{1pt}%
  \colorbox{paramyellow}{\strut #1}%
  \endgroup
}

\begin{table*}[ht]
\centering

\begin{subtable}[t]{0.43\textwidth}
\centering
\renewcommand{\arraystretch}{1.25}
\small
\resizebox{\linewidth}{!}{%
\begin{tabular}{p{1.8cm}p{4.7cm}}
\toprule
\textbf{Performance Level} & \textbf{Feedback} \\
\midrule
Low \newline (Reward $\leq$ 0) 
& \raggedright ``Your score is $-5$ points. Try to improve paddle positioning to \feedbackhl{prevent opponent scoring}.'' \tabularnewline
\midrule
Medium \newline(0 $<$ Reward $<$ 19) 
& \raggedright ``Keep it up! You're scoring \feedbackhl{12 points} against the opponent but you are still \feedbackhl{9 points} from winning the game. Try improving paddle positioning to \feedbackhl{prevent opponent scoring}.'' \tabularnewline
\midrule
High \newline(Reward $\geq$ 19) 
& \raggedright ``Good job! You're close to winning the game! You're scoring \feedbackhl{20 points} against the opponent, only \feedbackhl{1 point} short of winning.'' \tabularnewline
\bottomrule
\end{tabular}%
}
\caption{Pong feedback.}
\label{tab:atari-pong-feedback-structure}
\end{subtable}
\hfill
\begin{subtable}[t]{0.55\textwidth}
\centering
\renewcommand{\arraystretch}{1.25}
\small
\resizebox{\linewidth}{!}{%
\begin{tabular}{p{2.0cm}p{6.0cm}}
\toprule
\textbf{Performance Level} & \textbf{Feedback} \\
\midrule
Low \newline (Reward $\leq$ 0) 
& \raggedright ``Your score is -5 points. Try to improve paddle positioning to \feedbackhl{return the ball and avoid} \newline \feedbackhl{losing lives}.'' \tabularnewline
\midrule
Medium \newline(0 $<$ Reward $<$ 300) 
& \raggedright ``Keep it up! You're scoring \feedbackhl{50 points} against the opponent but you are still \feedbackhl{300 points} from winning the game. Try improving paddle positioning to \feedbackhl{return the ball and avoid losing} \newline \feedbackhl{lives}.'' \tabularnewline
\midrule
High \newline(Reward $\geq$ 300) 
& \raggedright ``Good job! You're close to winning the game! You're scoring \feedbackhl{320 points} against the opponent, \feedbackhl{try ensuring you return the ball}, only \feedbackhl{30 points} short of winning.'' \tabularnewline
\bottomrule
\end{tabular}%
}
\caption{Breakout feedback.}
\label{tab:feedback-structure-breakout}
\end{subtable}

\caption{Staged feedback templates for Pong and Breakout. Highlighted phrases mark the parts that differ between the corresponding High, Medium, and Low feedback levels across the two games.}
\label{tab:atari-feedback-structure}
\end{table*}

\begin{table}[ht]
\centering
\small
\renewcommand{\arraystretch}{1.3}
\begin{tabular}{p{6.5cm}p{6.5cm}}
\toprule
\textbf{Performance Level} & \textbf{Feedback} \\
\midrule
High \newline(Reward $\geq$ 300) & \raggedright ``Great job! You're performing well with an \feedbackhl{average score of 320}. Try to improve your \newline \feedbackhl{shooting accuracy} and \feedbackhl{dodging}.'' \tabularnewline
\midrule
Medium \newline(100 $\leq$ Reward $<$ 300) & \raggedright ``Good progress! Your \feedbackhl{average score is 180}. \newline Focus on better timing for \feedbackhl{shooting} and \feedbackhl{avoiding enemy projectiles}.'' \tabularnewline
\midrule
Low \newline (Reward $<$ 100) & \raggedright ``Your \feedbackhl{average score is 70}. Try to improve your  \newline strategy for \feedbackhl{shooting aliens} and \feedbackhl{dodging} \newline \feedbackhl{projectiles}.'' \tabularnewline
\bottomrule
\end{tabular}
\vspace{2mm}
\caption{Staged feedback for the Space Invaders agent at different performance levels.}
\label{tab:feedback-structure-space}
\end{table}

\subsection{Credit Horizon and Rollout Design}
\label{sec:app:atari:horizon}

Both conditions start from the same modular agent initialization. After collecting a trace, we append a templated natural-language feedback string derived from the observed rewards.

The main paper compares one-step and multi-step credit horizons across all eight games. In the one-step condition, the optimizer receives a trace containing a single action and its immediate reward and updates the policy after every step. In the multi-step condition, the optimizer receives a longer rollout before proposing a revision. The per-game rollout lengths are summarized in Table~\ref{tab:atari-training-configs}; the representative games discussed below use 400 steps for Pong, 300 for Breakout, and 25 for Space Invaders.

These values were chosen to balance two opposing forces. Longer traces reveal delayed consequences and are more faithful to the eventual control objective, but they also consume more of the optimizer's context budget and reduce update frequency. The representative games differ meaningfully in how informative local rewards are. Pong and Breakout provide immediate geometric cues that often align with full-episode success, whereas Space Invaders requires more strategic coordination between movement, shooting, and threat avoidance. This is why the credit-horizon choice is exposed as a design decision rather than left implicit in the implementation.

\paragraph{Credit-horizon comparison} Figure~\ref{fig:atari-learning-curve} shows that the credit horizon can be task-dependent. Under the aggregate comparison shown in the figure, multi-step optimization outperforms one-step optimization in four of the eight games (Pong, Breakout, Space Invaders, and Asterix), while one-step optimization performs better in the other four (Freeway, Enduro, Q*bert, and Seaquest). This split suggests that credit horizon is a genuine design choice.

The representative games in the appendix help interpret why the outcomes differ. In Space Invaders, a possible benefit of longer traces is that effective play requires coordinating shooting and movement under delayed consequences, whereas games such as Freeway can benefit from more frequent updates based on short-horizon feedback. Paddle-and-ball games such as Pong and Breakout sit between these extremes: they still admit relatively interpretable local geometric cues, but longer traces can remain useful because action quality depends on how returns shape future trajectories.

\subsection{Object-Centric Deep RL Baselines}
\label{sec:app:deep-rl-result}

The DQN and PPO curves in Figure~\ref{fig:atari-training-efficiency} were produced with object-centric variants of CleanRL~\citep{huang2022cleanrl}. We adapted an existing library rather than building a custom deep RL stack from scratch. In these runs, a master's student in Computer Science implemented the object-centric baselines in roughly three days, and the subsequent training, debugging, and result collection took about two weeks. The goal was to obtain reasonable neural baselines with modest engineering effort, not to carry out an exhaustive baseline-optimization campaign.

All deep RL agents used object-centric inputs from OCAtari~\citep{delfosse2024ocatari} rather than raw pixels. For the neural baselines, the object information was converted into numeric features and fed to multilayer perceptrons, rather than passed as the raw Python dictionary used by the LLM agent. The main implementation files were the shared object wrapper \texttt{obj\_atari\_env.py}\footnote{\anonurl{https://anonymous.4open.science/r/cleanrl_obj_centric-822B/cleanrl/obj_atari_env.py}{https://github.com/ameliakuang/cleanrl_obj_centric/blob/master/cleanrl/obj_atari_env.py}} together with the training scripts \texttt{obj\_dqn\_atari.py}\footnote{\anonurl{https://anonymous.4open.science/r/cleanrl_obj_centric-822B/cleanrl/obj_dqn_atari.py}{https://github.com/ameliakuang/cleanrl_obj_centric/blob/master/cleanrl/obj_dqn_atari.py}} and \texttt{obj\_ppo\_cleanrl.py}.\footnote{\anonurl{https://anonymous.4open.science/r/cleanrl_obj_centric-822B/cleanrl/obj_ppo_cleanrl.py}{https://github.com/ameliakuang/cleanrl_obj_centric/blob/master/cleanrl/obj_ppo_cleanrl.py}} Both inherited most algorithmic hyperparameters from the corresponding CleanRL implementations. We did not run a broad hyperparameter sweep or a systematic search over network families. The architecture changes we made were limited to a few hand-chosen MLP size adjustments that varied by game; for example, we used a smaller DQN encoder for Pong and larger ones for harder games such as Breakout and Space Invaders. Each run used 10M environment steps, and the experiment launch scripts ran five seeds per game.

\begin{table*}[t]
\centering

\begin{subtable}[t]{0.48\textwidth}
\centering
\renewcommand{\arraystretch}{1.15}
\small
\resizebox{\linewidth}{!}{%
\begin{tabular}{@{}p{2.7cm}p{2.1cm}p{2.1cm}@{}}
\toprule
\textbf{Setting} & \textbf{DQN baseline} & \textbf{PPO baseline} \\
\midrule
Observation
& \multicolumn{2}{p{4.2cm}}{
\shortstack[l]{OCAtari objects converted\\to normalized numeric vectors}
} \\
\midrule
Training steps per run & 10M & 10M \\
Parallel envs per job & 1 & 10 \\
Seeds per game & 5 & 5 \\
Batch size & 512 & 1280 \\
Minibatch size & --- & 320 \\
\bottomrule
\end{tabular}%
}
\caption{Concrete settings used for the object-centric deep RL baselines.}
\label{tab:atari-deeprl-setup}
\end{subtable}
\hspace{0.5em}
\begin{subtable}[t]{0.46\textwidth}
\centering
\renewcommand{\arraystretch}{1.15}
\small
\resizebox{\linewidth}{!}{%
\begin{tabular}{@{}p{2.3cm}p{2.2cm}p{2.2cm}@{}}
\toprule
\textbf{Algorithm} & \multicolumn{1}{c}{\textbf{Time (min)}} & \multicolumn{1}{c}{\textbf{Score (\%)}} \\ 
\midrule
Neural Net (DQN)
& \vtop{\hbox to 2.2cm{\hfil 291.6\hfil}\hbox to 2.2cm{\hfil (184.3--469.1)\hfil}\strut}
& \vtop{\hbox to 2.2cm{\hfil 71.5\hfil}\hbox to 2.2cm{\hfil (47.5--114.2)\hfil}\strut} \\ 
\midrule
Neural Net (PPO)
& \vtop{\hbox to 2.2cm{\hfil 219.3\hfil}\hbox to 2.2cm{\hfil (163.3--255.2)\hfil}\strut}
& \vtop{\hbox to 2.2cm{\hfil 108.1\hfil}\hbox to 2.2cm{\hfil (71.6--143.7)\hfil}\strut} \\ 
\midrule
Code (LLM)
& \vtop{\hbox to 2.2cm{\hfil 8.3\hfil}\hbox to 2.2cm{\hfil (4.3--21.0)\hfil}\strut}
& \vtop{\hbox to 2.2cm{\hfil 44.3\hfil}\hbox to 2.2cm{\hfil (3.0--100.2)\hfil}\strut} \\
\bottomrule
\end{tabular}%
}
\caption{Median runtime and score statistics, reported with interquartile range (25\%--75\%).}
\label{tab:median-stats}
\end{subtable}

\caption{Left: concrete settings for the object-centric deep RL baselines. Right: median runtime and score statistics across games.}
\label{tab:atari-setup-and-stats}
\end{table*}

All runs were executed on a single machine with 48 CPUs, 96 logical CPUs, and 8 NVIDIA A100-SXM4-40GB GPUs. Although the machine had 8 GPUs, each individual training job used only one 40GB A100. The remaining GPUs were used to run other jobs concurrently, not to accelerate a single run with multi-GPU training. Reported wall-clock times therefore reflect straightforward single-job, single-GPU training. 

Figure~\ref{fig:atari-training-efficiency} therefore compares against practical object-centric deep RL baselines implemented with limited tuning, rather than against the fastest achievable DQN or PPO systems. We use this to represent what a single developer can achieve in a reasonable amount of time. Instead of using the total training time, which is the duration of 10M steps, we use the logged timestamps to identify when each run first reaches its \textbf{highest score}. Since deep RL training is not stable and performance can regress instead of monotonically improve, this captures the time until the best solution is discovered, making the reporting protocol for deep RL results more comparable to that of LLM-based runs.

\subsection{Representative Learned Artifact Examples}
\label{sec:app:atari-code}

We provide representative initial code (with docstrings) together with the learned revisions for three games: Pong in Figures~\ref{fig:atari-pong-init-agent} and~\ref{fig:atari-pong-final-agent}, Breakout in Figures~\ref{fig:atari-breakout-init-agent-1}, \ref{fig:atari-breakout-init-agent-2}, \ref{fig:atari-breakout-final-agent-1}, and~\ref{fig:atari-breakout-final-agent-2}, and Space Invaders in Figures~\ref{fig:atari-space-invaders-init-agent-1}, \ref{fig:atari-space-invaders-init-agent-2}, and~\ref{fig:atari-space-invaders-final-agent-1}. Figure~\ref{fig:atari-strategy-snippets} highlights a few of the higher-level strategies that emerged in the learned code. These examples are intended to make the broader credit-horizon discussion concrete by showing the kinds of code policy the optimizer actually discovers.

\begin{figure*}[ht]
\centering
\begin{subfigure}[t]{0.32\linewidth}
\small
\begin{lstlisting}[language=Python]
# Increase margin and add dynamic adjustment
# based on ball distance
base_margin = 4
ball_x = obs['Ball'].get('x', 0)
dist_factor = (140 - ball_x) / 140
margin = base_margin * (1 + dist_factor)

# Add momentum-based adjustment
if obs['Ball'].get('dx', 0) > 0:
    ball_dy = obs['Ball'].get('dy', 0)
    predicted_ball_y += ball_dy * dist_factor
\end{lstlisting}
\caption{Pong: adaptive interception instead of pure ball chasing.}
\end{subfigure}\hfill
\begin{subfigure}[t]{0.32\linewidth}
\small
\begin{lstlisting}[language=Python]
# Ball in upper half - aim for tunnels
# to high bricks
if ball['y'] < 120:
    # Look for gaps in brick rows to target
    for color in ['RB', 'OB']:
        ...

    # Adjust paddle to deflect ball
    # toward high-value bricks
    if ball['x'] < high_brick_x:
        target_x = pre_ball_x - 4
\end{lstlisting}
\caption{Breakout: tunnel-seeking returns toward higher-value bricks.}
\end{subfigure}\hfill
\begin{subfigure}[t]{0.32\linewidth}
\small
\begin{lstlisting}[language=Python]
# There can only be one player bullet
# on the field at a time
for key, obj in obs.items():
    if key.startswith('Bullet') and obj.get('dy', 0) < 0:
        return False

# Move away from threats; otherwise move
# toward more aliens
if threat_left > threat_right:
    move = 1
elif aliens_right > aliens_left:
    move = 1
\end{lstlisting}
\caption{Space Invaders: respect firing constraints while balancing attack and defense.}
\end{subfigure}
\caption{\textbf{Representative high-level strategies discovered by the optimizer.} These short excerpts show that the learned revisions do more than tune local action thresholds: Pong adds distance- and momentum-aware interception, Breakout begins to target tunnels and higher-value bricks, and Space Invaders combines action constraints with threat-aware positioning.}
\label{fig:atari-strategy-snippets}
\end{figure*}

Figure~\ref{fig:atari-strategy-snippets} and the full code examples in Figures~\ref{fig:atari-pong-init-agent}--\ref{fig:atari-space-invaders-final-agent-1} suggest that the LLM optimizer often discovers compact game-specific heuristics rather than generic code cleanups. In Pong, the learned changes become more anticipatory; in Breakout, they begin to encode targeting decisions about where the return should send the ball next; and in Space Invaders, they coordinate firing and movement under environment constraints. These representative cases make the learned code policy easier to inspect and help explain why longer credit horizons can be especially useful in games with more strategic needs.

\setcounter{figure}{0}
\renewcommand{\thefigure}{C\arabic{figure}}
\setcounter{table}{0}
\renewcommand{\thetable}{C\arabic{table}}

\section{BigBench Extra Hard Details}
\label{sec:app:mini-batch-learning}

\subsection{Experimental Setup}
\label{sec:app:bbeh_setup}

BigBench Extra Hard (BBEH)~\citep{kazemi2025big} is a benchmark of challenging language-understanding tasks. We use eight tasks spanning logical reasoning (Dyck Languages, Boolean Expressions), spatial reasoning (Geometric Shapes), language understanding (Linguini, Disambiguation QA), recommendation and rule-based reasoning (Movie Recommendation, Boardgame QA), and causal reasoning (Causal Understanding).

\begin{table}[ht]
\centering
\begin{tabular}{lc}
\toprule
\textbf{Parameter} & \textbf{Value} \\
\midrule
Training examples per task & First 15 examples \\
Validation examples per task & Next 10 examples \\
Held-out test examples per task & Remaining examples (typically 175+) \\
\midrule
Batch sizes compared & 1, 3, 5 \\
Total optimizer update steps & 15 \\
Number of trials (seeds) & 3 \\
\midrule
LLM optimizer & OptoPrime \\
LLM backend & Claude Sonnet-3.5-v2 \\
\bottomrule
\end{tabular}
\vspace{2mm}
\caption{BigBench Extra Hard experimental configurations.}
\label{tab:bbeh-configs}
\end{table}

For each task, we form a fixed split from dataset order: the first 15 examples are used for training, the next 10 for validation, and the remainder for held-out testing. During optimization, we still perform random minibatching within the 15-example training set. At each optimizer update, we sample $k$ training examples, where $k \in \{1, 3, 5\}$ is the batch size under study, execute the current agent on each example, collect correctness feedback, and concatenate the resulting traces before sending them to the optimizer. Note that the 3 trials represent \textbf{different random batch shuffling of the same training set}.

\subsection{Agent Design Details}
\label{sec:app:bbeh_agent}

The training code is available online\footnote{\ifanonurl Anonymized\else \url{https://github.com/AgentOpt/OpenTrace/blob/experimental/examples/bbeh/bbeh_trace.py}\fi}.
It instantiates a simple two-part agent, matching the abstraction in the main paper. One part is a trainable prompt template that is concatenated with each question before calling the base LLM. The other is a trainable answer-extraction function that postprocesses the raw response into the exact output format expected by the evaluator.

\begin{figure}[ht]
    \centering
    \includegraphics[width=0.9\linewidth]{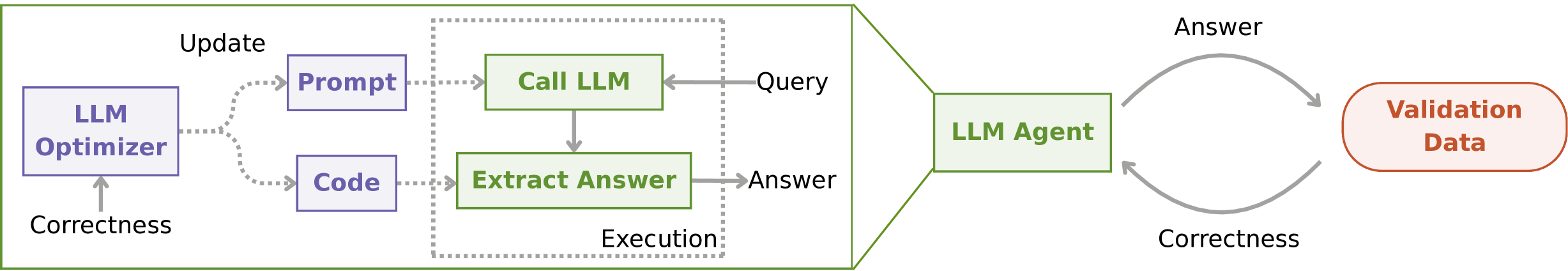}
    \caption{\textbf{Agent design for BigBench Extra Hard}. The agent consists of two optimizable components: \texttt{call\_llm} and \texttt{answer\_extraction}. During optimization, multiple traces can be concatenated through the batchify operator $\oplus$ before being shown to the optimizer.}
    \label{fig:bbeh-agent-graph}
\end{figure} 

\paragraph{\texttt{call\_llm}} In code, this corresponds to combining the trainable prompt template with the task query and passing the resulting string to the backend LLM. The prompt is the main optimization target and is revised across updates.

\paragraph{\texttt{answer\_extraction}} The trainable \texttt{extract\_answer} function parses the raw LLM response into the final answer. The initialized implementation simply splits on the string ``Answer:'', which makes formatting part of the optimization problem rather than assuming it is solved in advance.

The optimizer therefore receives traces containing the input query, the prompt used, the raw response, the extracted answer, and the resulting correctness feedback. This makes BBEH a clean testbed for the experience-batching question in the main paper because the editable artifact is small, but still includes both prompting and postprocessing decisions.

\subsection{Feedback Design Details}
\label{sec:app:bbeh_feedback}

Feedback is provided by a task-agnostic guide that returns a binary score together with a short natural-language message. When the prediction is correct, the guide returns a success message; when it is incorrect, the guide reveals the expected answer and asks for a revision to the prompt or program.

The metric is also intentionally simple. For multiple-choice tasks, the evaluator extracts the last answer of the form \texttt{(A)}, \texttt{(B)}, etc.; otherwise it compares the predicted string to the target by exact match. This deliberately minimal design lets us study batch size without introducing task-specific reward shaping.

\subsection{Batchify Operator Details}
\label{sec:app:batchify}

The batchify operator $\oplus$ concatenates execution traces from multiple independent examples into a single learning context. For batch size $k$, we construct
\[
\text{BatchTrace} = \text{Trace}_1 \oplus \text{Trace}_2 \oplus \cdots \oplus \text{Trace}_k.
\]

Each trace contains the question, the agent's predicted answer, and the feedback string. The concatenated batch trace is then passed to the optimizer as one context so that it can search for revisions that generalize across several examples rather than specializing to a single case.

The implementation keeps the total update budget fixed at 15 across batch sizes. Since each task has 15 training examples, this is realized by changing the number of passes over the training set: batch size 1 uses 1 epoch, batch size 3 uses 3 epochs, and batch size 5 uses 5 epochs. In all cases, the optimizer is allowed to make 15 updates per trial.

\subsection{Task-Specific Observations}
\label{sec:app:bbeh_observations}

\paragraph{Held-out test performance} Table~\ref{tab:structure_lang_tasks} reports accuracy across batch sizes. Batch size 1 achieves the highest test accuracy for Disambiguation QA (0.537) and Movie Recommendation (0.889), size 3 for Geometric Shapes (0.389), Linguini (0.234), and Boolean Expressions (0.238), and size 5 for Dyck Languages (0.190) and Causal Understanding (0.531).

\paragraph{Task-dependent batch size} The validation curves in Figure~\ref{fig:minibatch-learning-curve-full} show that the effect of batching is strongly task-dependent. Larger batches often produce smoother early learning, but they do not consistently yield the best final test performance. Smaller batches are noisier, yet sometimes continue improving for longer.

\subsection{Additional Framework Training Details}
\label{sec:app:bbeh_dspy_langgraph}

\begin{table}[htbp]
\centering
\caption{\textbf{Cross-framework check.} Test accuracy across minibatch sizes on the three shared tasks. Gray values give the standard deviation over 3 trials.}
\vspace{1mm}
\scriptsize
\setlength{\tabcolsep}{3pt}
\renewcommand{\arraystretch}{1.06}
\begin{tabular}{lcccc}
\toprule
Framework & Batch & Boardgame & Causal & Movie \\
\midrule
\multirow{3}{*}{LangGraph}
& 1 & 0.208 {\scriptsize \textcolor{gray}{$\pm$ 0.152}} 
    & 0.425 {\scriptsize \textcolor{gray}{$\pm$ 0.082}} 
    & 0.680 {\scriptsize \textcolor{gray}{$\pm$ 0.031}} \\
& 3 & 0.278 {\scriptsize \textcolor{gray}{$\pm$ 0.012}} 
    & 0.430 {\scriptsize \textcolor{gray}{$\pm$ 0.109}} 
    & 0.703 {\scriptsize \textcolor{gray}{$\pm$ 0.035}} \\
& 5 & \textbf{0.280} {\scriptsize \textcolor{gray}{$\pm$ 0.020}} 
    & \textbf{0.493} {\scriptsize \textcolor{gray}{$\pm$ 0.139}} 
    & \textbf{0.737} {\scriptsize \textcolor{gray}{$\pm$ 0.028}} \\
\midrule
\multirow{3}{*}{DSPy}
& 1 & 0.288 {\scriptsize \textcolor{gray}{$\pm$ 0.035}} 
    & \textbf{0.472} {\scriptsize \textcolor{gray}{$\pm$ 0.027}} 
    & 0.448 {\scriptsize \textcolor{gray}{$\pm$ 0.351}} \\
& 3 & \textbf{0.377} {\scriptsize \textcolor{gray}{$\pm$ 0.025}} 
    & 0.341 {\scriptsize \textcolor{gray}{$\pm$ 0.031}} 
    & 0.189 {\scriptsize \textcolor{gray}{$\pm$ 0.251}} \\
& 5 & 0.299 {\scriptsize \textcolor{gray}{$\pm$ 0.035}} 
    & 0.278 {\scriptsize \textcolor{gray}{$\pm$ 0.015}} 
    & \textbf{0.495} {\scriptsize \textcolor{gray}{$\pm$ 0.173}} \\
\bottomrule
\end{tabular}
\label{tab:cross_framework_wrap}
\end{table}

LangGraph obtains its highest test accuracy with batch size 5 on all three shared tasks, whereas the best batch size for DSPy varies across tasks (Table~\ref{tab:cross_framework_wrap}).

For the DSPy study, we implemented a minimal question-answering program using 
\texttt{dspy.Predict("question -> answer")} and optimized only the instruction. 
Since standard DSPy COPRO optimizer does not expose the same minibatch-style learning context as our Trace experiments, we implemented a batch-aware COPRO variant. 
At each optimization step, the current program is run on a minibatch of training examples, and we construct a feedback context containing, the question, ground-truth answer, model answer, and correctness, for each example. 
Then the batch-aware COPRO optimizer proposes a revised instruction from the current instruction and the concatenated feedback. 
We run 15 optimization steps for each batch size and select the best intermediate program using the 10-example validation set before reporting accuracy on the held-out test set.

For the LangGraph study, we implemented the same task as a small two-node graph: one node generates a response from a trainable prompt, and a second node extracts the final answer in the format expected by the evaluator. 
We then optimize the graph with OptoPrime by treating the prompt and graph functions as trainable parameters because LangGraph does not provide an optimizer. 
At each update step, the graph is executed on a minibatch of training examples, each prediction is converted into correctness feedback against the ground truth, and the feedback strings are concatenated into a single optimization context before the optimizer update. 
As in the main experiments, we use 15 training examples, 10 validation examples, and the remaining examples for testing, run 15 optimization steps, and select the best candidate by validation accuracy. 

Both studies use the same answer-matching rule as the main experiments: multiple-choice answers are matched by the predicted option token, while non-multiple-choice answers are matched exactly.

\setcounter{figure}{0}
\renewcommand{\thefigure}{D\arabic{figure}}
\setcounter{table}{0}
\renewcommand{\thetable}{D\arabic{table}}

\section{Harness Optimization Details}
\label{sec:app:harness}

\subsection{Three-Factor Sweep}
\label{sec:app:harness:sweep}

\paragraph{Protocol} Following work on optimizing model harnesses~\citep{lee2026metaharness}, the optimizer rewrites the harness source using a direct model call. We select harness checkpoints using the 15 TBLite validation tasks and reserve Terminal-Bench~2.0 for transfer. The single-file version places the agent logic in one function; the modular version separates prompts, output parsing, action policy, and the driver. All code is editable in both cases. With the full batch and full horizon, the context would reach roughly 1.1 million tokens. We truncate the end of the last trajectory to fit within the one-million-token limit.

\paragraph{Evaluation and sensitivity ranges} Each sweep score averages one run per task over a fixed subset of 15 of the 50 TBLite test tasks or 15 of the 89 Terminal-Bench~2.0 tasks. For Table~\ref{tab:harness-factor-sensitivity}, we vary one factor while fixing the other two at the single-file, batch-5, 50\%-horizon setting. The starting-artifact range uses the single-file, Mini-SWE-Agent, and Terminus-2 results in Table~\ref{tab:harness-starting-artifacts}. The modular result evaluates the seed and is excluded from this range and the optimized-harness grid. The horizon and batch-size ranges use the corresponding three settings in Table~\ref{tab:harness-grid}.

\begin{table}[htbp]
\centering
\small
\caption{Starting-artifact comparison at batch size 5 and a 50\% horizon. The modular seed score is shown separately from the optimized-harness results.}
\label{tab:harness-starting-artifacts}
\begin{tabular}{@{}lrr@{}}
\toprule
Starting artifact & TBLite & Terminal-Bench~2.0 \\
\midrule
Single-file & 0.800 & 0.267 \\
Mini-SWE-Agent & 0.333 & 0.333 \\
Terminus-2 & 0.800 & 0.400 \\
\midrule
Modular (seed) & 0.600 & 0.200 \\
\bottomrule
\end{tabular}
\end{table}

\begin{table}[htbp]
\centering
\small
\caption{Reported Terminal-Bench~2.0 scores for harnesses using Claude Haiku~4.5. Our row is the modular harness from Table~\ref{tab:harness-grid} with batch size five and a full credit horizon, evaluated on the 15-task transfer subset. The other scores are from the leaderboard at the time of the comparison.}
\vspace{1mm}
\label{tab:harness-tb2-comparison}
\begin{tabular}{@{}lr@{}}
\toprule
Harness & Terminal-Bench~2.0 score \\
\midrule
Optimized modular harness (ours) & 40.0\% \\
Goose & 35.5\% \\
Mini-SWE-Agent & 29.8\% \\
Terminus-2 & 28.3\% \\
\bottomrule
\end{tabular}
\end{table}

\subsection{Fixed-Budget Sweeps}
\label{sec:app:harness:budget}

\paragraph{Task splits} The separate TBLite budget sweep uses 40 training, 10 validation, and 50 test tasks. The DeepSWE sweep uses 40 training, 20 validation, and 53 test tasks. In both sweeps, we select the checkpoint with the highest validation score for each configuration and evaluate that checkpoint on the test set. These runs reuse the starting artifact that performed best in the TBLite study.

\begin{table}[htbp]
\centering
\small
\caption{Complete fixed-budget sweeps. Each budget has three tested combinations of batch size $B$ and horizon $L$ in steps. Bold marks the highest test score at each budget, including ties.}
\vspace{1mm}
\label{tab:harness-budget-sweeps}
\begin{tabular}[t]{@{}rrrc@{}}
\toprule
\multicolumn{4}{c}{TBLite} \\
\midrule
Budget & $B$ & $L$ & Test \\
\midrule
15k & 1 & 44 & 0.459 \\
    & 3 & 6 & 0.467 \\
    & 6 & 2 & \textbf{0.470} \\
\addlinespace
26k & 2 & 44 & 0.319 \\
    & 5 & 8 & \textbf{0.528} \\
    & 12 & 2 & 0.474 \\
\addlinespace
65k & 6 & 31 & \textbf{0.626} \\
    & 13 & 8 & 0.441 \\
    & 31 & 2 & 0.357 \\
\addlinespace
123k & 11 & 44 & 0.469 \\
     & 21 & 10 & \textbf{0.479} \\
     & 40 & 4 & 0.390 \\
\addlinespace
247k & 22 & 44 & 0.505 \\
     & 30 & 19 & 0.522 \\
     & 40 & 11 & \textbf{0.545} \\
\bottomrule
\end{tabular}
\qquad
\begin{tabular}[t]{@{}rrrcc@{}}
\toprule
\multicolumn{5}{c}{DeepSWE} \\
\midrule
Budget & $B$ & $L$ & Validation & Test \\
\midrule
60k & 1 & 160 & 0.324 & 0.199 \\
    & 2 & 56 & 0.270 & 0.229 \\
    & 4 & 24 & 0.263 & \textbf{0.237} \\
\addlinespace
100k & 2 & 110 & 0.257 & 0.186 \\
     & 4 & 48 & 0.237 & \textbf{0.251} \\
     & 8 & 20 & 0.313 & 0.231 \\
\addlinespace
170k & 3 & 150 & 0.183 & 0.217 \\
     & 6 & 56 & 0.212 & \textbf{0.230} \\
     & 14 & 20 & 0.214 & \textbf{0.230} \\
\addlinespace
285k & 5 & 160 & 0.173 & \textbf{0.224} \\
     & 11 & 52 & 0.251 & 0.216 \\
     & 23 & 20 & 0.232 & 0.170 \\
\addlinespace
480k & 9 & 130 & 0.211 & 0.237 \\
     & 19 & 48 & 0.218 & \textbf{0.267} \\
     & 40 & 20 & 0.230 & 0.186 \\
\bottomrule
\end{tabular}
\end{table}

\paragraph{Calibrating on either domain} Calibrating the horizon on DeepSWE gives a fraction of $52/76$, or a target of about eight steps on TBLite. In both cases, we apply the fraction to the target domain's pilot estimate of $\rho$ and choose the tested horizon nearest to that target. Table~\ref{tab:deep-swe-horizon-transfer} reports those choices. For each strategy in Table~\ref{tab:harness-budget-regret}, we subtract the selected test score from the best test score among the three configurations at the same budget and average over budgets.

\setcounter{figure}{0}
\renewcommand{\thefigure}{E\arabic{figure}}
\setcounter{table}{0}
\renewcommand{\thetable}{E\arabic{table}}

\section{Building a Learning Loop: Extended Discussion}
\label{sec:algorithm}

A learning loop can be built in many ways. Here, we discuss some of the key concepts in a learning loop by presenting the optimization problem using a graph formalism.
This particular viewpoint is not commonly used in practice, but is the backbone of TextGrad~\citep{yuksekgonul2024textgrad} and DSPy~\citep{khattab2023dspy}.
At a high level, the discussion below is analogous to how TensorFlow or PyTorch combine per-example or per-timestep computation into a larger optimization graph, for example, through \texttt{reduce\_mean} / \texttt{sum}, concatenation operators such as \texttt{cat} / \texttt{stack}, or temporal unrolling across a sequence like \texttt{scan}.

We first discuss the intuitive notions of a \emph{learning loop} and a \emph{learning context} using the framework of OPTO (Optimization with Trace Oracle). 
We then connect that formalism back to the engineering decisions highlighted in the main paper: how the workflow is modularized, which components are made optimizable, and what form of feedback is supplied to the optimizer.

We note that there are many other forms of learning, such as building an explicit memory of past experiences (i.e., a generate-retrieve-summarize system)~\citep{zhou2025mem1,ouyang2025reasoningbank}, or updating LLM parameters during test time (i.e., test-time training)~\citep{yuksekgonul2026learning}.
We focus only on learning through the lens of optimization, where agent learning happens during an optimization step, after the agent has interacted with the environment and collected sufficient experiences.

\vspace{1em}

\subsection{The OPTO Framework}
\label{sec:app:formalizing_learning_loop}

OPTO (Optimization with Trace Oracle)~\citep{cheng2024trace} was proposed as a unified framework for describing iterative generative optimization problems. An OPTO problem (a generalization of numerical optimization) is described by a tuple $(\Theta, \omega, \mathcal{T})$, where $\Theta$ is the parameter space, $\omega$ is the problem context and $\mathcal{T}$ is a Trace Oracle. For a parameter $\theta \in \Theta$, the Trace Oracle $\mathcal{T}$ returns a tuple $(f, g)$ where $g$ is a computational graph involving $\theta$ and $f$ is a feedback signal provided to exactly one node in $g$ -- the output node.

\subsection{Workflow Graphs and Learning Graphs}
\label{sec:app:workflow_graphs}

\paragraph{Example.} We define a workflow $W_\theta$ with two functions $h_{\theta_1}, h_{\theta_2}$, each controlled by its own parameter. Given an input $x_i$, let the intermediate node be $o_i = h_{\theta_1}(x_i)$ and the output be $y_i = h_{\theta_2}(o_i)$. A single execution of the workflow on $x_i$ induces a graph $g_i$ with directional edges $(x_i, \theta_1) \rightarrow o_i$ and $(o_i, \theta_2) \rightarrow y_i$ (see Figure~\ref{fig:graph-g}), together with feedback $f_i$ attached to the output node. We refer to $g_i$ as the \emph{workflow graph} produced by this execution. In other words, $W_\theta$ denotes the parameterized system, while $g_i$ denotes one concrete execution trace of that system.

This already captures many workflows. For example, for a customer-support agent, $h_{\theta_1}$ can be an RAG database query function, where $\theta_1$ is the retrieval search query. Once the database returns a list of items, $h_{\theta_2}$ can be an LLM call that synthesizes those items into a final answer, and $\theta_2$ represents the tunable instruction for the answer style and content.

\begin{figure}[ht!]
\centering
\resizebox{0.45\textwidth}{!}{%
\begin{tikzpicture}[
    node distance=1.2cm and 1.0cm,
    every node/.style={font=\small},
    var/.style={circle, draw, minimum size=0.7cm, inner sep=0pt, align=center},
    param/.style={rectangle, draw, rounded corners, minimum width=0.6cm, align=center},
    ->, >=Stealth
]

\node[var] (x) {$x$};
\node[var, right=of x, draw=paramblueEdge, fill=paramblue] (o) {$o$};
\node[var, right=of o] (y) {$y$};
\node[var, right=of y, draw=feedbackpurpleEdge, fill=feedbackpurple] (f) {$f$};

\node[param, below left=0.3cm and 0.4cm of o, draw=paramyellowEdge, fill=paramyellow] (t1) {$\theta_1$};
\node[param, below left=0.3cm and 0.4cm of y, draw=paramyellowEdge, fill=paramyellow] (t2) {$\theta_2$};

\draw (x) -- (o);
\draw (o) -- (y);
\draw[dashed] (f) -- (y);
\draw (t1) -- (o);
\draw (t2) -- (y);

\end{tikzpicture}
}%
\caption{An example workflow graph $g_i$ produced by a single execution of $W_\theta$. The graph contains input node $x_i$, intermediate node $o_i$, output node $y_i$, feedback $f_i$, and the parameter nodes $\theta_1$, $\theta_2$ that influence the output.}
\label{fig:graph-g}
\end{figure}

The workflow graph $g_i$ can be viewed as one \emph{experience} of the system: one input, one execution, one output, and its associated feedback. This is often enough for single-instance optimization, but it is not enough to specify many learning problems of interest. An agent may need to generalize across multiple independent examples, or delay its update until a full sequence of interactions has finished. In such cases, the optimizer should not receive only one workflow graph; it should receive a larger graph that represents the right unit of learning.

We call this optimizer-facing object the \emph{learning graph}, denoted by $G_{\mathrm{learn}}$. It is constructed from one or more workflow graphs through a fixed \emph{learning template} $T$:
\[
G_{\mathrm{learn}} = T(g_1, \dots, g_k).
\]
The template determines how individual experiences are combined before being shown to the optimizer. In the language of OPTO, $G_{\mathrm{learn}}$ is the graph returned to the optimizer for the learning problem under consideration.

\begin{figure}
    \centering
    \includegraphics[width=\linewidth]{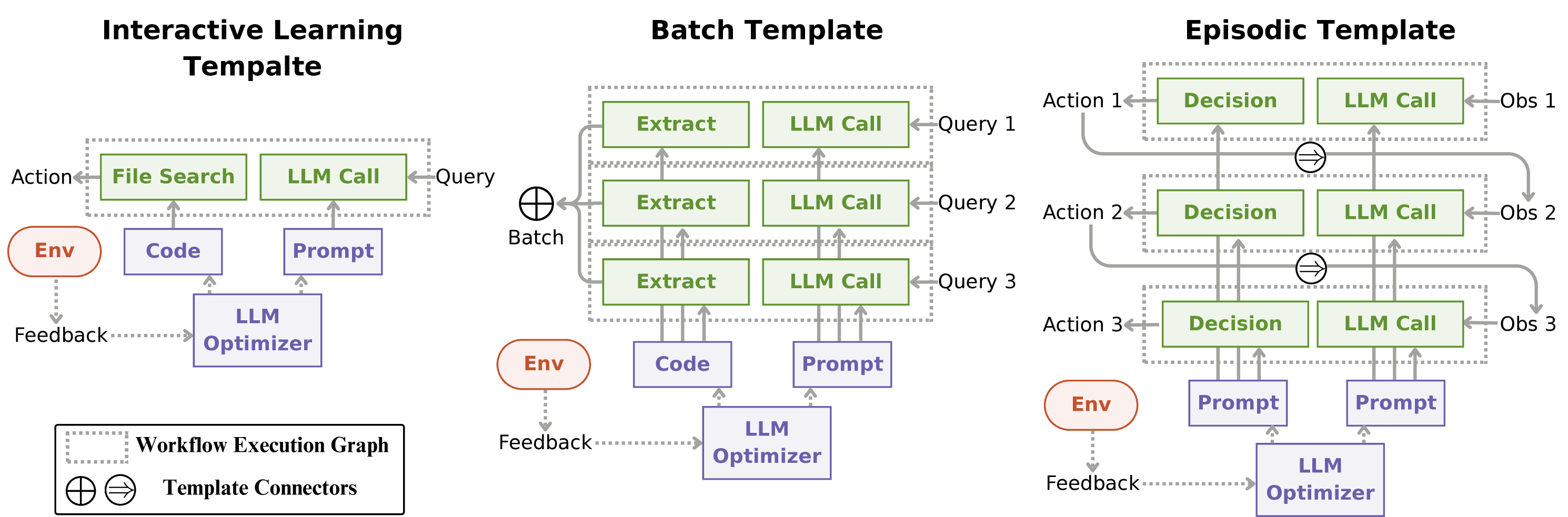}
    \caption{\textbf{Representing different learning problems through graph templates.} Each dotted rectangle denotes a workflow graph produced by a single execution of the workflow. A learning template specifies how one or more workflow graphs are combined into the learning graph $G_{\mathrm{learn}}$ shown to the optimizer. In batch learning, independent workflow graphs are aggregated using $\oplus$. In episodic learning, workflow graphs are linked through temporal/environment transitions denoted by $\Rightarrow$ and the result is presented to the optimizer.}
    \label{fig:learning-graph}
    \vspace{-4mm}
\end{figure}

\subsection{Building Learning Graphs by Template}
\label{sec:app:graph_templates}

We now describe how to build the learning graph for common learning problems. We assume a workflow $W_\theta$ is given, where $\theta$ denotes the optimizable parameters. Without loss of generality, the workflow takes an input $x_i$ and returns an output $y_i = W_\theta(x_i)$, although both $x_i$ and $y_i$ may be structured objects rather than scalars. The workflow here may represent an entire software system, including multiple LLM calls, tools, retrieval steps, and decision rules. A single execution of this workflow on $x_i$ produces a workflow graph $g_i$.

We further assume a feedback oracle that maps an execution into feedback $f_i$, which can be numerical, textual, visual, or structured. We do not assume that the feedback is differentiable or even scalar; we only assume it contains useful information about the quality of the execution. Given a collection of workflow graphs $\{g_i\}$ and associated feedback $\{f_i\}$, a learning template specifies how they should be combined into $G_{\mathrm{learn}}$ before being passed to the generative optimizer.

\textbf{Interactive Learning Template.} Here the agent learns on the fly as it interacts with the world~\citep{shalev2012online}. At each step, it sees an input $x_1$, outputs $y_1$, receives feedback $f_1$, and then updates its parameter $\theta$. No cross-example composition is needed:
\[
G_{\mathrm{learn}} = T_{\mathrm{interactive}}(g_1) = g_1.
\]
This covers online-learning and bandit-style settings where a single experience already defines one update.

\textbf{Batch Learning Template.} In batch learning, the agent should update from several \emph{independent} examples at once~\citep{hastie2009elements}. Consider a minibatch $\big[(x_i, z_i) \big]_{i=1}^B$, where $z_i$ denotes the task-specific supervision or comparison signal associated with input $x_i$. Executing the workflow on each example yields workflow graphs $\{g_i\}_{i=1}^B$ with outputs $\{o_i\}_{i=1}^B$ and feedback $\{f_i\}_{i=1}^B$. The batch template aggregates these experiences into one learning graph:
\[
\begin{aligned}
G_{\mathrm{learn}} &= T_{\mathrm{batch}}(g_1, \dots, g_B), \\
\hat{o} &= \bigoplus_{i=1}^B o_i.
\end{aligned}
\]
The aggregated feedback $\bigoplus_{i=1}^B f_i$ is then attached to $\hat{o}$. Intuitively, the batchify operator $\oplus$ plays the role of concatenating multiple experiences into one optimizer-facing context.

\textbf{Episodic Learning Template.} In episodic learning, the experiences are not independent. Instead, the output of one execution influences the next observation through the environment. We therefore link the workflow graphs sequentially:
\[
g_1 \Rightarrow g_2 \Rightarrow \cdots \Rightarrow g_T,
\]
where $\Rightarrow$ denotes the environment transition from one step to the next. This is the key difference from batch learning: the examples are chained causally rather than merely aggregated. After constructing this trajectory-level structure, we obtain the learning graph
\[
G_{\mathrm{learn}} = T_{\mathrm{episodic}}(g_1, \dots, g_T),
\]
which can optionally include aggregated episode-level observations or feedback, such as $\hat{o} = \bigoplus_{i=1}^T o_i$.

\textbf{Remark.} Choosing the right learning template matters because it determines the learning context shown to the optimizer. If we want a parameter that generalizes across a dataset but instead optimize from one example at a time, the optimizer may become overly sensitive to presentation order. Conversely, if an agent's actions have delayed consequences, then changing the behavior before the episode terminates can create objective mismatch. The distinction between workflow graphs, learning templates, and the resulting learning graph makes explicit what counts as one update in the learning loop.

\subsection{The Choice of Starting Artifact}
\label{sec:app:starting_artifact}

The starting artifact is not a single object, but a set of design decisions about what the LLM optimizer is allowed to edit and what prior structure it starts from. 
Different agentic frameworks expose different interfaces for these choices. 
The OPTO framework naturally maps the starting artifact to three concrete components. 
This decomposition is useful because it makes clear what inductive bias is being given to the LLM optimizer.

\paragraph{Workflow Structure and Specs.} First, the engineer chooses the number of files, functions, or program components that make up the workflow, together with how they connect to one another. In OPTO, this appears as the graph connectivity and the input-output signature of each function. In other frameworks, the same choice may appear as interface specifications between files, tool schemas, or a plan for how different modules call each other. This choice matters because it determines the search space: changing one monolithic function is a different optimization problem from changing a set of coordinated components with explicit interfaces.
However, this is usually overlooked in many agentic frameworks, or is mixed together with program documentation.

\paragraph{Program Documentation.} Second, the engineer can provide docstrings, comments, instructions, or other textual descriptions that explain what each function or program component is intended to do. In OPTO, function docstrings are a natural way to instantiate this idea. In other agentic frameworks, the same role may be fulfilled by documents, instruction manuals, and READMEs. 
These artifacts shape the optimizer's interpretation of the system and therefore can be changed by engineers to guide the optimizer's search.

\paragraph{Initial Implementations.} Third, the engineer can decide how much starter code to provide before optimization begins. 
At one extreme, a function may be left almost entirely unspecified except for its signature and documentation. 
At the other extreme, the engineer may provide a partial or even fully working implementation and ask the optimizer to improve or refactor it. 
In all cases, the amount of initial implementation changes what the LLMs see as the starting context and changes the initial experience, i.e., whether the program succeeds or fails.

\subsection{The Choice of Credit Horizon}
\label{sec:app:credit_horizon}

Besides deciding how the workflow graph is constructed, the rest of the design choices are about how the learning graph is constructed.
The learning graph is the context that is provided as input to the optimizer.
If the applicable learning template is intended to capture the long-term effects of the workflow, then the credit horizon is the number of steps to include in the learning graph.

The choice of credit horizon is commonly studied in RL, though instead of directly deciding a horizon length to truncate, usually it is dynamically balanced by a discount factor~\citep{badia2020agent57}.
Direct credit-horizon truncation is, however, more commonly studied in recurrent neural networks, where truncated back-propagation through time (TBPTT) is used to truncate the gradient computation at a fixed horizon~\citep{pascanu2013difficulty,tallec2017unbiasing,shaban2019truncated}.

\subsection{The Choice of Experience Batching}
\label{sec:app:experience_batching}

When a batch learning template is used, the engineer has decided that a workflow must generalize to and accommodate a diverse set of inputs and perform well to handle them.
This dictates that the LLM optimizer that is used to update the workflow cannot overfit to just a single example -- it must learn to reason across different examples sampled from a larger distribution.
Thus, experience batching closely mimics the study of batch size in stochastic gradient descent, where the number of examples aggregated per update affects both learning dynamics and generalization~\citep{smith2017don}.
It is also closely related to active learning, where examples are selected to be put in a batch to make learning more efficient~\citep{houlsby2011bayesian,gal2017deep}, as well as contrastive learning, where positive and negative examples are balanced in a batched setting~\citep{chen2020simple,doumbouya2025tversky}.

\subsection{Additional Considerations: Feedback Design}
\label{sec:app:feedback_design}

Feedback design is an important part of the learning loop, but we do not attempt to treat it comprehensively in this paper. We instead point readers to prior work that explores informative and effective feedback design~\citep{chen2023teaching,nie2024importance,wei2024improving,xu2025provably}. In the case-study appendices below, we mainly use staged and suggestive feedback and provide concrete examples of how it is instantiated.

\textbf{Staged and Suggestive Feedback.} The simplest feedback reports correctness or reward. In practice, however, the engineer often has additional freedom to vary the feedback according to performance regimes or execution states. The staged reward templates in the Atari appendix and the templates in the MLAgentBench appendix are examples of this design choice.

\setcounter{figure}{0}
\renewcommand{\thefigure}{F\arabic{figure}}
\setcounter{table}{0}
\renewcommand{\thetable}{F\arabic{table}}

\section{Further Discussion and Disclosures}
\label{sec:app:discussion}

\subsection{Agents, Agentic Systems, and the Scope of This Paper}
\label{sec:app:agents_scope}

\begin{figure*}
    \centering
    \includegraphics[width=\linewidth]{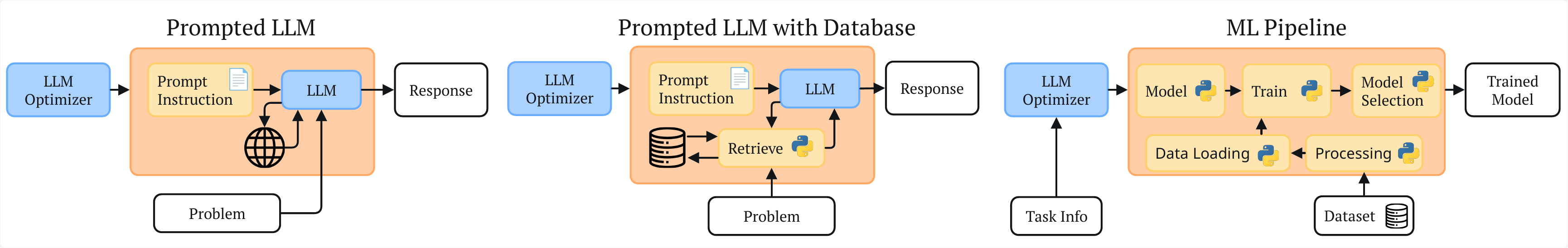}
    \caption{Examples of LLM-based generative optimization applied to different kinds of systems: a prompted system with a search tool (\textit{left}), a more complex prompting setup with retrieval code (\textit{middle}), and an end-to-end machine learning pipeline (\textit{right}). We use {\color{paramblue}\rule{1em}{0.6em}} to mark an LLM API call and {\color{paramyellow}\rule{1em}{0.6em}} to mark text or code files.}
    \label{fig:agent-examples}
\end{figure*}

There is ongoing disagreement in the literature about how narrowly to define an ``LLM agent.'' One useful definition is that an LLM agent is a system in which the LLM repeatedly selects actions through interaction with an environment and receives feedback from that environment. Under this view, some systems optimized in this paper are plainly agents, while others are better understood as software artifacts that \emph{contain} LLM components.

For the purposes of this paper, the distinction is less important than the shared optimization structure. Whether the target is a prompt, a retrieval routine, a game-playing policy, or a multi-function software workflow (see Figure~\ref{fig:agent-examples}), the engineering problem is to specify an initial artifact, execute it, collect feedback, and present the resulting evidence to a generative optimizer. This shared structure is what motivates our use of the term \emph{learning loop}.

\begin{figure*}
    \centering
    \includegraphics[width=0.9\linewidth]{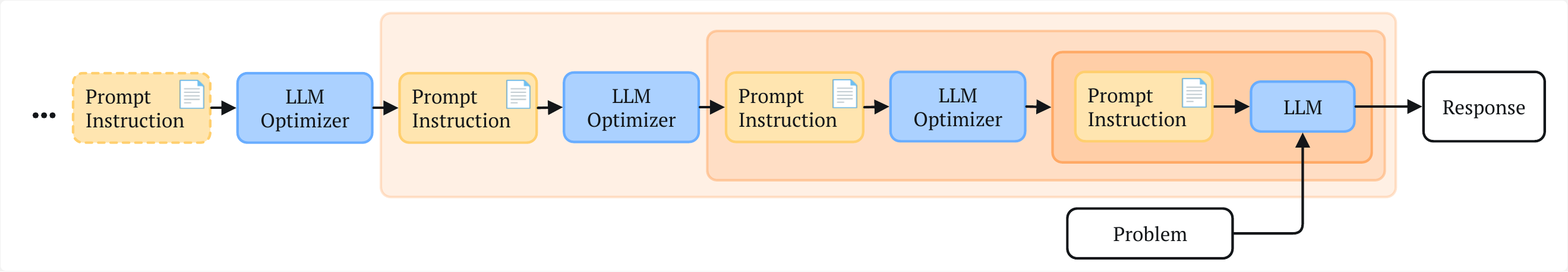}
    \caption{\textbf{Infinite nested agentic system optimization}. It is easy to construct an example in which one optimizer writes the instructions for another optimizer, creating an infinite regress of optimization problems. We do not study that setting in this paper and instead focus on commonly used, finite agentic systems.}
    \label{fig:infinitely_nested}
\end{figure*}

Figure~\ref{fig:infinitely_nested} illustrates an interesting case of the learning-by-optimization perspective: we can always find another layer to optimize -- we can construct a recursively nested optimization problem where an optimizer optimizes the prompts or instructions of an inner optimizer. Such examples are conceptually interesting, but they obscure the more practically relevant design issues studied in the main paper. The appendix therefore uses them only to clarify the scope of our claims.

\subsection{Disclosures and LLM Access Cards}
\label{sec:app:disclosures}

\subsubsection{Large Language Model Access Card}
\label{sec:app:llm_access_card}

The experiments in Section~\ref{sec:factors} were conducted during the period of February 2025 to April 2025. The primary model used for optimization in those experiments was Anthropic Claude Sonnet-3.5-v2. Additional harness-optimization experiments in Section~\ref{sec:harness-experiments} were conducted in 2026, using Claude Sonnet~5 to rewrite the harness and Claude Haiku~4.5 as the agent model. Whenever we compare against other learned-agent baselines, including re-run implementations of prior methods, we use the same model endpoint during the same period whenever possible so that differences in access conditions do not dominate the comparison.

\subsubsection{Large Language Model Use for Writing}
\label{sec:app:llm_writing}

A small number of sentences and paragraphs in the paper were polished with GPT-5 after the authors had already drafted the original text. The model was used for editing and polishing rather than for generating new technical claims, results, or arguments from scratch. 

\subsection{Existing Assets, Licenses, and Terms of Use}
\label{sec:app:licenses}

We use only public benchmark datasets, public simulation environments, open-source libraries, and closed-source LLM APIs accessed through provider endpoints. The main software assets are: Trace/OpenTrace (MIT License), MLAgentBench (MIT License), OCAtari (MIT License), Gymnasium (MIT License), Arcade Learning Environment / ale-py (GPL-2.0), CleanRL and our object-centric CleanRL fork (MIT License, preserving upstream notices), DSPy (MIT License), and LangGraph (MIT License). The main benchmark/data assets are: BigBench Extra Hard (BBEH; Apache License 2.0 for the repository, with several benchmark task directories marked Creative Commons Attribution 4.0 International), BIG-bench where applicable (Apache License 2.0), Kaggle House Prices -- Advanced Regression Techniques (MIT License), and Kaggle Spaceship Titanic (Creative Commons Attribution 4.0 International). Kaggle data are not redistributed in our code release; readers are directed to obtain the data from Kaggle after accepting the relevant competition rules and Kaggle terms. 

\setcounter{figure}{0}
\renewcommand{\thefigure}{G\arabic{figure}}
\setcounter{table}{0}
\renewcommand{\thetable}{G\arabic{table}}

\clearpage
\section{Optimized Code Examples}
\label{sec:app:code_listings}

The large code examples referenced in the appendices are grouped here so that figures and tables in the earlier appendix sections can stay closer to the corresponding discussion.

\subsection{MLAgentBench Code Examples}
\label{sec:app:code_listings:mlagent}

\begingroup
\renewenvironment{figure}[1][]%
  {\par\medskip\captionsetup{type=figure}}%
  {\par\medskip}

\begin{figure}[ht]
    \centering
\begin{lstlisting}[language=Python]
import trace

@trace.model
class SpaceshipTitanicPipeline(Module):

    @trace.bundle(trainable=True)
    def preprocess(self, data):
        """
        [...docstring is skipped to save space...]
        """
        # Create a copy to avoid modifying original data
        # Handle missing values in numeric columns
        numeric_columns = ["Age", "RoomService", "FoodCourt", "ShoppingMall",
            "Spa", "VRDeck"]
        for col in numeric_columns:
            processed_data[col] = processed_data[col].fillna(processed_data[col].median())
    
        # Handle boolean/categorical columns
        processed_data["VIP"] = processed_data["VIP"].fillna(False)
        processed_data["CryoSleep"] = processed_data["CryoSleep"].fillna(False)
    
        # Convert HomePlanet to numeric using label encoding
        if "HomePlanet" in processed_data.columns:
            processed_data["HomePlanet"] = processed_data["HomePlanet"].fillna("Unknown")
            planet_map = {"Earth": 0, "Europa": 1, "Mars": 2, "Unknown": 3}
            processed_data["HomePlanet"] = processed_data["HomePlanet"].map(planet_map)

        # (skipped some code)
        
        # Age-related features
        processed_data["Age"] = processed_data["Age"].fillna(processed_data["Age"].median())
        processed_data["AgeGroup"] = pd.qcut(
            processed_data["Age"], q=6, labels=[0, 1, 2, 3, 4, 5]
        ).astype(int)

        # Interaction features
        processed_data["CryoSleepVIP"] = processed_data["CryoSleep"].astype(int) * processed_data["VIP"].astype(int)
        processed_data["SpendingPerAge"] = processed_data["TotalSpending"] / processed_data["Age"].clip(lower=1)
        processed_data["HasSpent"] = (processed_data["TotalSpending"] > 0).astype(int)
        processed_data["SpendingVariety"] = (processed_data[spending_columns] > 0).sum(axis=1)

        # ... standard scaling, dropping columns, etc.
        
        # Final check for NaN values
        processed_data = processed_data.fillna(0)
        return processed_data
\end{lstlisting}
\caption{Final code for Spaceship-Titanic Learned Agent. Docstrings are generated by ChatGPT and then edited by humans.}
\label{fig:mlagent-spaceship-final-agent-1}
\end{figure}

\subsection{Atari Code Examples}
\label{sec:app:code_listings:atari}

\begin{figure}[ht]
    \centering
\begin{lstlisting}[language=Python]
import trace

@trace.model
class Policy(Module):
    def __call__(self, obs):
        predicted_ball_y = self.predict_ball_trajectory(obs)
        action = self.select_action(predicted_ball_y, obs)
        return action

    @trace.bundle(trainable=True)
    def predict_ball_trajectory(self, obs):
        """
        Predict the y-coordinate where the ball will intersect with the player's paddle by calculating its trajectory,
        using ball's (x, y) and (dx, dy) and accounting for bounces off the top and bottom walls.

        Game Setup:
        - Screen dimensions: The game screen has boundaries where the ball bounces
          - Top boundary: approximately y=30
          - Bottom boundary: approximately y=190
        - Paddle positions:
          - Player paddle: right side of screen (x = 140)
          - Enemy paddle: left side of screen (x = 16)

        Args:
            obs (dict): Dictionary containing object states for "Player", "Ball", and "Enemy".
                       Each object has position (x,y), size (w,h), and velocity (dx,dy).

        Returns:
            float: Predicted y-coordinate where the ball will intersect the player's paddle plane.
                  Returns None if ball position cannot be determined.

        """
        if 'Ball' in obs:
            return obs['Ball'].get("y", None)
        return None
    
    @trace.bundle(trainable=True)
    def select_action(self, predicted_ball_y, obs):
        '''
        Select the optimal action to move player paddle by comparing current player position and predicted_ball_y.
        
        IMPORTANT! Movement Logic:
        - If the player paddle's y position is GREATER than predicted_ball_y: Move DOWN (action 2)
          (because the paddle needs to move downward to meet the ball)
        - If the player paddle's y position is LESS than predicted_ball_y: Move UP (action 3)
          (because the paddle needs to move upward to meet the ball)
        - If the player paddle is already aligned with predicted_ball_y: NOOP (action 0)
          (to stabilize the paddle when it's in position)
        Ensure stable movement to avoid missing the ball when close by.

        Args:
            predicted_ball_y (float): predicted y coordinate of the ball or None
            obs(dict): Dictionary of current game state, mapping keys ("Player", "Ball", "Enemy") to values (dictionary of keys ('x', 'y', 'w', 'h', 'dx', 'dy') to integer values)
        Returns:
            int: 0 for NOOP, 2 for DOWN, 3 for UP
        '''

        if predicted_ball_y is not None and 'Player' in obs:
            return random.choice([2, 3])
        return 0
\end{lstlisting}
\caption{Initial code for Pong agent.}
\label{fig:atari-pong-init-agent}
\end{figure}

\begin{figure}[ht]
    \centering
\begin{lstlisting}[language=Python]
import trace

@trace.model
class Policy(Module):
    def __call__(self, obs):
        predicted_ball_y = self.predict_ball_trajectory(obs)
        action = self.select_action(predicted_ball_y, obs)
        return action

    @trace.bundle(trainable=True)
    def predict_ball_trajectory(self, obs):
        """(same as before, skipped to save space)"""
        if 'Ball' in obs:
            ball = obs['Ball']
            # If ball moving away from player, return None
            if ball.get('dx', 0) < 0:
                return None
                
            # Calculate time to reach paddle
            paddle_x = 140
            ball_x = ball.get('x', 0)
            ball_dx = ball.get('dx', 0)
            if ball_dx == 0:
                return ball.get('y', None)
                
            time_to_paddle = (paddle_x - ball_x) / ball_dx
            
            # Calculate predicted y position with improved accuracy
            ball_y = ball.get('y', 0)
            ball_dy = ball.get('dy', 0)
            predicted_y = ball_y + ball_dy * time_to_paddle
            
            # Account for bounces with improved accuracy
            num_bounces = 0
            while predicted_y < 30 or predicted_y > 190:
                if predicted_y < 30:
                    predicted_y = 30 + (30 - predicted_y)
                if predicted_y > 190:
                    predicted_y = 190 - (predicted_y - 190)
                num_bounces += 1
                if num_bounces > 4:  # Limit bounce calculations
                    break
                    
            return predicted_y
        return None
    
    @trace.bundle(trainable=True)
    def select_action(self, predicted_ball_y, obs):
        '''(same as before, skipped to save space)'''
        if predicted_ball_y is not None and 'Player' in obs:
            # Calculate center of paddle
            paddle_center = obs['Player']['y'] + obs['Player']['h']/2
            
            # Increase margin and add dynamic adjustment based on ball distance
            base_margin = 4
            if 'Ball' in obs:
                ball_x = obs['Ball'].get('x', 0)
                dist_factor = (140 - ball_x) / 140  # Normalized distance factor
                margin = base_margin * (1 + dist_factor)  # Larger margin when ball is far
                
                # Add momentum-based adjustment
                if obs['Ball'].get('dx', 0) > 0:
                    ball_dy = obs['Ball'].get('dy', 0)
                    # Scale adjustment based on distance
                    predicted_ball_y += ball_dy * dist_factor
            else:
                margin = base_margin
            
            # More aggressive movement thresholds
            if paddle_center > predicted_ball_y + margin:
                return 2  # Move down
            elif paddle_center < predicted_ball_y - margin:
                return 3  # Move up
            return 0  # Stay in position
        return 0
\end{lstlisting}
\caption{Final learned code for Pong agent.}
\label{fig:atari-pong-final-agent}
\end{figure}

\begin{figure}[ht]
    \centering
\begin{lstlisting}[language=Python]
@trace.model
class Policy(Module):
    def __call__(self, obs):
        pre_ball_x = self.predict_ball_trajectory(obs)
        target_paddle_pos = self.generate_paddle_target(pre_ball_x, obs)
        action = self.select_paddle_action(target_paddle_pos, obs)
        return action

    @trace.bundle(trainable=True)
    def predict_ball_trajectory(self, obs):
        """
        Predict the x-coordinate where the ball will intersect with the player's paddle by calculating its trajectory,
        using ball's (x, y) and (dx, dy) and accounting for bounces off the right and left walls.

        Game setup: 
        - Screen dimensions: The game screen has left and right walls and brick wall where the ball bounces 
          - Left wall: x=9
          - Right wall: x=152
        - Paddle positions:
          - Player paddle: bottom of screen (y=189)
        - Ball speed:
          - Ball deflects from higher-scoring bricks would have a higher speed and is harder to catch.
        - The paddle would deflect the ball at different angles depending on where the ball lands on the paddle
        
        Args:
            obs (dict): Dictionary containing object states for "Player", "Ball", and blocks "{color}B" (color in [R/O/Y/G/A/B]).
                       Each object has position (x,y), size (w,h), and velocity (dx,dy).
        Returns:
            float: Predicted x-coordinate where the ball will intersect the player's paddle plane.
                  Returns None if ball position cannot be determined.
        """
        if 'Ball' not in obs:
            return None
            
    @trace.bundle(trainable=True)
    def generate_paddle_target(self, pre_ball_x, obs):
        """
        Calculate the optimal x coordinate to move the paddle to catch the ball (at predicted_ball_x)
        and deflect the ball to hit bricks with higher scores in the brick wall.

        Logic:
        - Prioritize returning the ball when the ball is coming down (positive dy)
        - The brick wall consists of 6 vertically stacked rows from top to bottom:
          - Row 1 (top): Red bricks (7 pts)
          - Row 2: Orange (7 pts)
          - Row 3: Yellow (4 pts)
          - Row 4: Green (4 pts)
          - Row 5: Aqua (1 pt)
          - Row 6 (bottom): Blue (1 pt)
         - Strategic considerations:
          - Breaking lower bricks can create paths to reach higher-value bricks above
          - Creating vertical tunnels through the brick wall is valuable as it allows
            the ball to reach and bounce between high-scoring bricks at the top
          - Balance between safely returning the ball and creating/utilizing tunnels
            to access high-value bricks
        - Ball speed increases when hitting higher bricks, making it harder to catch

        Args:
            pre_ball_x (float): predicted x coordinate of the ball intersecting with the paddle or None
            obs (dict): Dictionary containing object states for "Player", "Ball", and blocks "{color}B" (color in [R/O/Y/G/A/B]).
                       Each object has position (x,y), size (w,h), and velocity (dx,dy).
        Returns:
            float: Predicted x-coordinate to move the paddle to. 
                Returns None if ball position cannot be determined.
        """
        if pre_ball_x is None or 'Ball' not in obs:
            return None
        return None
\end{lstlisting}
\caption{Initial code for Breakout agent (Part 1).}
\label{fig:atari-breakout-init-agent-1}
\end{figure}

\begin{figure}[ht]
    \centering
\begin{lstlisting}[language=Python]
import trace

@trace.model
class Policy(Module):

    # (continued from above)

    @trace.bundle(trainable=True)
    def select_paddle_action(self, target_paddle_pos, obs):
        """
        Select the optimal action to move player paddle by comparing current player position and target_paddle_pos.

        Movement Logic:
        - If the player paddle's center position is GREATER than target_paddle_pos: Move LEFT (action 3)
        - If the player paddle's center position is LESS than target_paddle_pos: Move RIGHT (action 2)
        - If the player paddle is already aligned with target_paddle_pos: NOOP (action 0)
          (to stabilize the paddle when it's in position)
        Ensure stable movement to avoid missing the ball when close by.

        Args:
            target_paddle_pos (float): predicted x coordinate of the position to best position the paddle to catch the ball,
                and hit the ball to break brick wall.
            obs (dict): Dictionary containing object states for "Player", "Ball", and blocks "{color}B" (color in [R/O/Y/G/A/B]).
                Each object has position (x,y), size (w,h), and velocity (dx,dy).
        Returns:
            int: 0 for NOOP, 2 for RIGHT, 3 for LEFT
        """
        if target_paddle_pos is None or 'Player' not in obs:
            return 0
            
        paddle = obs['Player']
        paddle_x = paddle['x']
        paddle_w = paddle['w']
        paddle_center = paddle_x + (paddle_w / 2)
        
        # Add deadzone to avoid oscillation
        deadzone = 2
        if abs(paddle_center - target_paddle_pos) < deadzone:
            return 0  # NOOP if close enough
        elif paddle_center > target_paddle_pos:
            return 3  # LEFT
        else:
            return 2  # RIGHT
\end{lstlisting}
\caption{Initial code for Breakout agent (Part 2).}
\label{fig:atari-breakout-init-agent-2}
\end{figure}

\begin{figure}[ht]
    \centering
\begin{lstlisting}[language=Python]
@trace.model
class Policy(Module):

    def __call__(self, obs):
        pre_ball_x = self.predict_ball_trajectory(obs)
        target_paddle_pos = self.generate_paddle_target(pre_ball_x, obs)
        action = self.select_paddle_action(target_paddle_pos, obs)
        return action

    @trace.bundle(trainable=True)
    def predict_ball_trajectory(self, obs):
        """(same as before, skipped to save space)"""
        if pre_ball_x is None or 'Ball' not in obs or 'Player' not in obs:
            return None
            
        ball = obs['Ball']
        paddle = obs['Player']
        
        # Default to centering paddle on predicted ball position
        target_x = pre_ball_x
        
        # Adjust paddle position based on current ball direction and brick locations
        if ball['dy'] > 0:  # Ball moving down
            if ball['y'] < 120:  # Ball in upper half - aim for tunnels to high bricks
                # Look for gaps in brick rows to target
                high_brick_x = None
                for color in ['RB', 'OB']:  # Check red and orange rows
                    if color in obs:
                        bricks = obs[color]
                        if len(bricks) > 0:
                            brick = bricks[0]
                            high_brick_x = brick['x'] + (brick['w'] / 2)
                            break
                            
                if high_brick_x is not None:
                    # Adjust paddle to deflect ball toward high-value bricks
                    if ball['x'] < high_brick_x:
                        target_x = pre_ball_x - 4  # Hit ball on right side
                    else:
                        target_x = pre_ball_x + 4  # Hit ball on left side
                        
        # Ensure target is within screen bounds
        paddle_half_width = paddle['w'] / 2
        target_x = max(9 + paddle_half_width, min(152 - paddle_half_width, target_x))
        
        return target_x
\end{lstlisting}
\caption{Learned code for Breakout agent (Part 1).}
\label{fig:atari-breakout-final-agent-1}
\end{figure}

\begin{figure}[ht]
    \centering
\begin{lstlisting}[language=Python]
@trace.model
class Policy(Module):

    # (continued from above)

    @trace.bundle(trainable=True)
    def generate_paddle_target(self, pre_ball_x, obs):
        """(same as before, skipped to save space)"""
        if 'Ball' not in obs:
            return None
            
        ball = obs['Ball']
        ball_x = ball['x']
        ball_y = ball['y']
        ball_dx = ball['dx']
        ball_dy = ball['dy']
        
        # If ball is not moving, can't predict
        if ball_dx == 0 and ball_dy == 0:
            return None
            
        # Calculate time to reach paddle
        paddle_y = 189  # Player paddle y-position
        if ball_dy != 0:  # Avoid division by zero
            time_to_paddle = (paddle_y - ball_y) / ball_dy
            if time_to_paddle > 0:  # Only predict if ball is moving toward paddle
                # Account for wall bounces
                x = ball_x + (ball_dx * time_to_paddle)
                num_bounces = 0
                while x < 9 or x > 152:
                    if x < 9:
                        x = 9 + (9 - x)  # Reflect off left wall
                    if x > 152:
                        x = 152 - (x - 152)  # Reflect off right wall
                    num_bounces += 1
                    if num_bounces > 10:  # Prevent infinite bounces
                        break
                return x
        return None

    @trace.bundle(trainable=True)
    def select_paddle_action(self, target_paddle_pos, obs):
        """(same as before, skipped to save space)"""
        if target_paddle_pos is None or 'Player' not in obs:
            return 0
            
        paddle = obs['Player']
        paddle_x = paddle['x']
        paddle_w = paddle['w']
        paddle_center = paddle_x + (paddle_w / 2)
        
        # Add deadzone to avoid oscillation
        deadzone = 2
        if abs(paddle_center - target_paddle_pos) < deadzone:
            return 0  # NOOP if close enough
        elif paddle_center > target_paddle_pos:
            return 3  # LEFT
        else:
            return 2  # RIGHT
        
\end{lstlisting}
\caption{Learned code for Breakout agent (Part 2).}
\label{fig:atari-breakout-final-agent-2}
\end{figure}

\begin{figure}[ht]
    \centering
\begin{lstlisting}[language=Python]
@trace.model
class Policy(Module):

    def __call__(self, obs):
        shoot_decision = self.decide_shoot(obs)
        move_decision = self.decide_movement(obs)
        return self.combine_actions(shoot_decision, move_decision)

    @trace.bundle(trainable=True)
    def decide_shoot(self, obs):
        '''
        Decide whether to shoot based on enemy positions and existing projectiles.
         
        Args:
            obs (dict): Game state observation containing object states for "Player", "Shield0", "Shield1", "Alien0", "Alien1", etc.
            Each object has position (x,y), size (w,h), and velocity (dx,dy).
            Player bullets have negative dy velocity and alien bullets have positive dy velocity
        
        Strategy tips:
        - You can only have one missile at a time
        - Try to shoot when aliens are aligned with your ship
        - Prioritize shooting at lower aliens as they're closer to you
        - Consider the movement of aliens when deciding to shoot
        
        Returns:
            bool: True if should shoot, False otherwise
        '''
        
        # There can only be one player bullet on the field at a time
        # Check for player bullets (which have negative dy velocity)
        for key, obj in obs.items():
            if key.startswith('Bullet') and obj.get('dy', 0) < 0:
                return False
            
        return random.choice([True, False])            

    @trace.bundle(trainable=True)
    def decide_movement(self, obs):
        '''
        Decide movement direction based on enemy positions and projectiles.
         
        Args:
            obs (dict): Game state observation containing object states for "Player", "Shield0", "Shield1", "Alien0", "Alien1", etc.
            Each object has position (x,y), size (w,h), and velocity (dx,dy).
            Player bullets have negative dy velocity and alien bullets have positive dy velocity
        
        Strategy tips:
        - Move to dodge enemy projectiles
        - Position yourself under aliens to shoot them
        - Stay away from the edges of the screen
        - Consider moving toward areas with more aliens to increase score
        
        Returns:
            int: -1 for left, 1 for right, 0 for no movement
        '''

        player = obs['Player']
        
        return random.choice([-1,0,1]) 

\end{lstlisting}
\caption{Initial code for Space Invaders agent (Part 1).}
\label{fig:atari-space-invaders-init-agent-1}
\end{figure}

\begin{figure}[ht]
    \centering
\begin{lstlisting}[language=Python]
@trace.model
class Policy(Module):

    # (continued from above)

    @trace.bundle(trainable=True)
    def combine_actions(self, shoot, movement):
        '''
        Combine shooting and movement decisions into final action.
        
        Args:
            shoot (bool): Whether to shoot
            movement (int): Movement direction
        
        Action mapping:
        - 0: NOOP (no operation)
        - 1: FIRE (shoot without moving)
        - 2: RIGHT (move right without shooting)
        - 3: LEFT (move left without shooting)
        - 4: RIGHT+FIRE (move right while shooting)
        - 5: LEFT+FIRE (move left while shooting)
        
        Returns:
            int: Final action (0: NOOP, 1: FIRE, 2: RIGHT, 3: LEFT, 4: RIGHT+FIRE, 5: LEFT+FIRE)
        '''
        
        if shoot and movement > 0:
            return 4  # RIGHT+FIRE
        elif shoot and movement < 0:
            return 5  # LEFT+FIRE
        elif shoot:
            return 1  # FIRE
        elif movement > 0:
            return 2  # RIGHT
        elif movement < 0:
            return 3  # LEFT
        return 0  # NOOP
\end{lstlisting}
\caption{Initial code for Space Invaders agent (Part 2).}
\label{fig:atari-space-invaders-init-agent-2}
\end{figure}

\begin{figure}[ht]
    \centering
\begin{lstlisting}[language=Python]
@trace.model
class Policy(Module):

    def __call__(self, obs):
        shoot_decision = self.decide_shoot(obs)
        move_decision = self.decide_movement(obs)
        return self.combine_actions(shoot_decision, move_decision)

    @trace.bundle(trainable=True)
    def decide_shoot(self, obs):
        """(same as before , skipped to save space)"""
        
        # There can only be one player bullet on the field at a time
        # Check for player bullets (which have negative dy velocity)
        for key, obj in obs.items():
            if key.startswith('Bullet') and obj.get('dy', 0) < 0:
                return False
            
        player = obs['Player']
        for key, obj in obs.items():
            if key.startswith('Alien'):
                # Check if alien is aligned with player (within 5 pixels)
                if abs(obj['x'] - player['x']) < 5:
                    # Prioritize lower aliens (higher y value)
                    if obj['y'] > 60:  # Adjust this threshold as needed
                        return True
        return False     

    @trace.bundle(trainable=True)
    def decide_movement(self, obs):
        """(same as before , skipped to save space)"""
        player = obs['Player']
        move = 0
        threat_left = 0
        threat_right = 0
        aliens_left = 0
        aliens_right = 0
        
        for key, obj in obs.items():
            if key.startswith('Alien'):
                if obj['x'] < player['x']:
                    aliens_left += 1
                else:
                    aliens_right += 1
            elif key.startswith('Bullet') and obj['dy'] > 0:  # Enemy bullet
                if obj['x'] < player['x']:
                    threat_left += 1
                else:
                    threat_right += 1
        
        # Move away from threats
        if threat_left > threat_right:
            move = 1
        elif threat_right > threat_left:
            move = -1
        # If no immediate threat, move towards more aliens
        elif aliens_left > aliens_right:
            move = -1
        elif aliens_right > aliens_left:
            move = 1
        
        return move    

    @trace.bundle(trainable=True)
    def combine_actions(self, shoot, movement):
        """(same as before , skipped to save space)"""
        if shoot and movement > 0:
            return 4  # RIGHT+FIRE
        elif shoot and movement < 0:
            return 5  # LEFT+FIRE
        elif shoot:
            return 1  # FIRE
        elif movement > 0:
            return 2  # RIGHT
        elif movement < 0:
            return 3  # LEFT
        return 0  # NOOP

\end{lstlisting}
\caption{Learned code for Space Invaders agent.}
\label{fig:atari-space-invaders-final-agent-1}
\end{figure}

\endgroup

\end{document}